# An Expressive Language and Efficient Execution System for Software Agents

**Greg Barish**                                                        gbarish@fetch.com
*Fetch Technologies*
*2041 Rosecrans Avenue, Suite 245*
*El Segundo, CA 90245 USA*

**Craig A. Knoblock**                                                  knoblock@isi.edu
*University of Southern California*
*Information Sciences Institute*
*4676 Admiralty Way*
*Marina del Rey, CA 90292 USA*

## Abstract

Software agents can be used to automate many of the tedious, time-consuming information processing tasks that humans currently have to complete manually. However, to do so, agent plans must be capable of representing the myriad of actions and control flows required to perform those tasks. In addition, since these tasks can require integrating multiple sources of remote information – typically, a slow, I/O-bound process – it is desirable to make execution as efficient as possible. To address both of these needs, we present a flexible software agent plan language and a highly parallel execution system that enable the efficient execution of expressive agent plans. The plan language allows complex tasks to be more easily expressed by providing a variety of operators for flexibly processing the data as well as supporting subplans (for modularity) and recursion (for indeterminate looping). The executor is based on a streaming dataflow model of execution to maximize the amount of operator and data parallelism possible at runtime. We have implemented both the language and executor in a system called THESEUS. Our results from testing THESEUS show that streaming dataflow execution can yield significant speedups over both traditional serial (von Neumann) as well as non-streaming dataflow-style execution that existing software and robot agent execution systems currently support. In addition, we show how plans written in the language we present can represent certain types of subtasks that cannot be accomplished using the languages supported by network query engines. Finally, we demonstrate that the increased expressivity of our plan language does not hamper performance; specifically, we show how data can be integrated from multiple remote sources just as efficiently using our architecture as is possible with a state-of-the-art streaming-dataflow network query engine.

## 1. Introduction

The goal of software agents is to automate tasks that require interacting with one or more accessible software systems. Past research has yielded several types of agents and agent frameworks capable of automating a wide range of tasks, including: processing sequences of operating system commands (Golden, Etzioni, & Weld, 1994), mediation of heterogeneous data sources (Wiederhold 1996; Bayardo, Bohrer, Brice, Cichocki, Fowler, Helal, Kashyap, Ksiezyk, Martin, Nodine, Rashid, Rusinkiewicz, Shea, Unnikrishnan, Unruh, & Woelk 1997; Knoblock, Minton, Ambite, Ashish, Muslea, & Tejada 2001), online comparison shopping (Doorenbos,

Etzioni, & Weld, 1996), continual financial portfolio analysis (Decker, Sycara, & Zeng, 1996), and airline ticket monitoring (Etzioni, Tuchinda, Knoblock, & Yates, 2004), to name only a few. Despite software agent heterogeneity, two recurring characteristics are (i) *the wide variety of tasks that agents are used to automate* and (ii) *the frequent need to process and route information during agent execution*.

Perhaps no other domain poses as many tantalizing possibilities for software agent automation as the Web. The ubiquity and practicality of the Web suggests that many potential benefits can be gained from automating tasks related to sources on the web. Furthermore, the Web is ripe for such automation – given the sheer number of online applications and the complete lack of coordination between them, agents could address an endless list of needs and problems to be solved for people that do use the Web for practical purposes. Furthermore, like other software agent domains, Web tasks vary widely in complexity and, by definition, involve routing and processing information as part of the task.

In this paper, we describe a software agent plan language and execution system that enables one to express a wide range of tasks as a software agent plan and then to have that plan be efficiently executed. We have implemented both the language and the executor in a system called THESEUS. Throughout this paper, we will discuss THESEUS in the context of Web information gathering and processing, since the Web represents a domain where most (if not all) of the challenges that software agents face can be found.

## 1.1 Web Information Agents

In recent years, the Web has experienced a rapid rate of growth, with more and more useful information becoming available online. Today, there exists an enormous amount of online data that people can not only view, but also use in order to accomplish real tasks. Hundreds of thousands of people use the Web every day to research airfares, monitor financial portfolios, and keep up to date with the latest news headlines. In addition to its enormity, what is compelling about the Internet as a practical tool is its dynamic, up-to-the-minute nature. For example, although information such as stock quotes and ticket availabilities change frequently, many sources on the Internet are capable of reporting these updates immediately. For this reason, and because of the breadth and depth of information it provides, the Web has become – for certain tasks – a more timely and necessary medium than even the daily newspaper, radio, or television.

The degree of complexity in gathering information from the Web varies significantly. Some types of tasks can be accomplished manually because the size of the data gathered is small or the need to query is infrequent. For example, finding the address of a restaurant or a theater in a particular city using a Yellow Pages type of Web site is easy enough for people to do themselves. It does not need to be automated, since the query need only be done once and the result returned is small and easy to manage. However, not all information gathering tasks are as simple. There are often times when the amount of data involved is large, or the answer requires integrating data from multiple sites, or the answer requires multiple queries over a period of time. For example, consider shopping for an expensive product over a period of time using multiple sources that are each updated daily. Such tasks can become quickly tedious and require a greater amount of manual work, making them very desirable to automate.

### 1.1.1. MORE COMPLICATED TASKS

One type of difficult Web information gathering task involves interleaved gathering and navigation. For the benefit of people that use a Web browser to access online data, many Web sources display large sets of query results spread over a series of web pages connected through "Next Page" links. For example, querying an online classified listings source for automobiles for sale can generate many results. Instead of displaying the results on a single very long Web page, many classified listings sites group sets of results over series of hyperlinked pages. In order to automatically collect this data, a system needs to interleave navigation and gathering an indeterminate number of times: that is, it needs to collect results from a given page, navigate to the next, gather the next set of results, navigate, and so on, until it reaches the end of set of results. While there has been some work addressing how to theoretically incorporate navigation into the gathering process (Friedman, Levy, & Millstein, 1999), no attention has been given to the efficient execution of plans that engage in this type of interleaved retrieval.

A second example has to do with monitoring a Web source. Since the Web does not contain a built-in trigger facility, one is forced to manually check sources for updated data. When updates are frequent or the need to identify an update immediately is urgent, it becomes desirable to automate the monitoring of these updates, notifying the user when one or more conditions are met. For example, suppose we want to be alerted as soon as a particular type of used car is listed for sale by one or more online classified ad sources. Repeated manual checking for such changes is obviously tedious. Mediators and network query engines can automate the query, but additional software in programming languages such as Java or C must be written to handle the monitoring process itself, something that requires conditional execution, comparison with past results, possible notification of the user, and other such actions.

### 1.1.2. THE NEED FOR FLEXIBILITY

These examples show that automatically querying and processing Web data can involve a number of subtasks, such as interleaved navigation and gathering and integration with local databases. Because of these needs, traditional database query languages like SQL are insufficient for the Web. The root of the problem is lack of flexibility, or expressivity, in these languages -- typically, only querying is supported. More complicated types of Web information gathering tasks, such as those described here and in other articles (Etzioni & Weld, 1994; Doorenbos et al., 1997; Chalupsky, Gil, Knoblock, Lerman, Oh, Pynadath, Russ, & Tambe, 2001; Ambite, Barish, Knoblock, Muslea, Oh, & Minton, 2002; Sycara, Paolucci, van Velsen, & Giampapa, 2003; Graham, Decker, & Mersic, 2003), usually involve actions beyond those needed for merely querying (i.e., beyond filtering and combining) – they require plans capable of a variety of actions, such as conditional execution, integration with local databases, and asynchronous notification to users. In short, Web information gathering tasks require an expressive query or plan language with which to describe a solution.

XQuery (Boag, Chamberlin, Fernandez, Florescu, Robie, & Simeon, 2002), used for querying XML documents, is one language that offers more flexibility. For example, XQuery supports "FLWOR" expressions that allow one to easily specify how to iterate over data. XQuery also supports conditional expressions, UDFs, and recursive functions.

Support for expressive agent plans can also be found in a number of software agent and robot agent frameworks, such as INFOSLEUTH (Bayardo et al., 1997), RETSINA (Sycara et al., 2003), DECAF (Graham et al., 2003), RAPs (Firby 1994) or PRS-LITE (Myers 1996). These systems support concurrent execution of operators and the ability to execute more complicated types of plans, such as those that require conditionals. In addition, unlike database systems, software agent and robot agent execution plans can contain many different types of operators, not just those limited to querying and filtering data.

### 1.1.3. THE NEED FOR EFFICIENCY

Despite support for more expressive plans, existing software agent and robot agent plan execution systems lack efficiency – a problem that is painfully realized when doing any kind of large scale data integration or when working with remote sources that operate at less than optimal speeds. In particular, while systems like RETSINA, DECAF, RAPs and PRS-LITE ensure a high-degree of *operator parallelism* (independent operators can execute concurrently), they do not ensure any type of *data parallelism* (independent elements of data can be processed concurrently). For example, it is not possible for one operator in these systems to *stream* information to another. This is understandable in the case of robot plan execution systems, which address how a robot interacts in the physical world, and are typically concerned with communicating effects, such as "has object X" or "direction north", which are – relatively speaking – small amounts of local information. Interestingly, while other software agent frameworks like DECAF and INFOSLEUTH have expressed a desire to support some sort of streaming architecture (Bayardo et al., 1997), such research has been constrained in part by the use of data transport layers (such as KQML) that do not contain the infrastructure necessary to support streaming.

However, for a software agent to process large amounts of remote information efficiently, both types of parallelism are critical. Dataflow-style parallelism is important in order to schedule independent operations concurrently; streaming is important in order to be able to process remote information as it becomes available and to make maximum use of local processing resources. Consider an agent that engages in two types of image processing on image data downloaded from a surveillance satellite. If it normally takes one minute to download the data and another minute for each type of processing, streaming dataflow execution can theoretically reduce the overall execution time by up to two-thirds. Even greater speedups are possible for different information processing tasks.

Though existing software agent and robot plan execution systems do not support streaming, a substantial amount of previous work has gone into building such architectures for database systems (Wilschut & Apers, 1993) and more recently network query engines (Ives, Florescu, Friedman, Levy, & Weld, 1999; Hellerstein, Franklin, Chandrasekaran., Deshpande, Hildrum, Madden, Raman, & Shah, 2000; Naughton, DeWitt, Maier, Aboulnaga, Chen, Galanis, Kang, Krishnamurthy, Luo, Prakash, Ramamurthy, Shanmugasundaram, Tian, Tufte, Viglas, Wang, Zhang, Jackson, Gupta, & Che, 2001). These systems employ special iterative-style operators that are aware of the underlying support for streaming[1] and exploit that feature to minimize operator blocking. Examples include the pipelined hash join (Wilschut & Apers, 1993; Ives, et al., 1999) the eddy data structure (Avnur & Hellerstein, 2000), which efficiently routes streaming data to operators. Despite support for a streaming dataflow model of execution, network query engines lack the generality and flexibility of existing agent frameworks and systems. XQuery does provide a more powerful and flexible language for Web data gathering and manipulation. However, the network query engines support only a subset of XQuery or related XML query processing operators and do not support constructs such as conditionals and recursion[2], which are essential for more complex types of information processing tasks.

**1.2 Contributions**

In summary, while it is desirable to automate the gathering and processing of data on the Web, it is currently not possible to build an agent that is both flexible and efficient using existing technologies. Existing agent execution systems are flexible in the types of plans they support, but they lack the ability to stream information, a critical feature that needs to be built into both the underlying architecture as well as the individual operators (i.e., operators need to implemented as iterators). Network query engines contain support for streaming dataflow, but lack the expressivity provided by existing agent plan languages.

In this paper, we address the need to combine both by presenting an expressive plan language and an efficient execution system for software agents. More specifically, this paper makes two contributions. The first is a software agent plan language that extends features of existing agent plan languages and is more expressive than the query languages of existing information integration systems and network query engines. The language proposed consists of a rich set of operators that, beyond gathering and manipulating data, support conditional execution, management of data in local persistent sources, asynchronous notification of results to users, integration between relational and XML sources, and extensibility (i.e., user-defined functions). In addition, the language is modular and encourages re-use: existing plans can be called from other plans as subplans and plans can be called recursively. Both the operators and constructs provide the expressivity necessary to address more complicated types of software agent tasks, such as the monitoring and interleaved navigation and gathering of Web data.

---

[1] Note: The term "pipelining" is common in database literature (e.g., Graefe, 1993), although "streaming" is often used in network query engine literature (see recent publications by Niagara and Telegraph).

[2] For example, Tukwila supports only a subset of Xquery (Ives et al., 2002).

A second contribution of this paper is the design for an executor that efficiently processes agent plans written in the proposed language. The core of the executor implements a streaming dataflow architecture, where data is dispatched to consuming operators as it becomes available and operators execute whenever possible. This design allows plans to realize the maximum degree of operational (horizontal) and data (vertical) parallelism possible. Our design also supports recursive streaming, resulting in the efficient execution of plans that require indeterminate looping, such as agents that interleave navigation and gathering of information on the Web. In short, the executor supports the highly parallel execution of software agent plans, leading to significant performance improvement over that provided by expressive, but less efficient agent executors.

We have implemented both the plan language and executor in a system called THESEUS. Throughout this paper, we refer to example plans that have been deployed in THESEUS in order to better illustrate plan expressivity and, later, to validate efficiency claims.

### 1.3 Organization

The rest of this paper is organized as follows. Section 2 provides background and the basic terminology of both dataflow computing (Arvind & Nikhil, 1990; Papdopoulos & Culler, 1990; Gurd & Snelling, 1992), generic information integration (Chawathe, Garcia-Molina, Hammer, Ireland, Papakonstantinou, Ullman, & Widom, 1994; Arens, Knoblock, & Shen, 1996; Levy, Rajaraman, & Ordille, 1996; Weiderhold 1996; Genesereth, Keller, & Duschka, 1997) and automated Web information gathering (Knoblock et al., 2001; Ives et al., 1999; Barish & Knoblock, 2002; Thakkar, Knoblock, & Ambite, 2003; Tuchinda & Knoblock, 2004). Section 3 describes the details involved in one type of complex software agent information gathering task, an example that will be used throughout the rest of the paper. In Section 4, we describe the proposed plan language in detail. Section 5 deals with the design of the streaming dataflow executor and how it provides high degrees of horizontal and vertical parallelism at runtime. In Section 6, we present experimental results, describing how the plan language and executor implemented in THESEUS measure up to those provided by other systems. In Section 7, we discuss the related work in greater detail. Finally, we present overall conclusions and discuss future work.

## 2. Preliminaries

The language and execution system we present in this paper build upon a foundation of prior research related to dataflow computing (Dennis 1974) and Web information integration (Wiederhold 1996; Bayardo et al., 1997, Knoblock et al., 2001). Although seemingly orthogonal disciplines, they are effective complements in that the parallelism and asynchrony provided by dataflow computing lends itself to the performance problems associated with Web information gathering.

### 2.1 Dataflow Computing

The pure **dataflow** model of computation was first introduced by Dennis (1974) as an alternative to the standard von Neumann execution model. Its foundations share much in common with past work on computation graphs (Karp & Miller, 1955), process networks (Kahn 1974), and communicating sequential processes (Hoare 1978). Dataflow computing has a long theoretical and experimental history, with the first machines being proposed in the early 1970s and real physical systems being constructed in the late 1970s and throughout the 1980s and early 1990s (Arvind & Nikhil, 1990; Papdopoulos & Culler, 1990; Gurd & Snelling, 1992).

The dataflow model of computation describes program execution in terms of data dependencies between instructions. A dataflow graph is a directed acyclic graph (DAG) of nodes and edges. The nodes are called **actors**. They consume and produce data **tokens** along the edges that connect them to other actors. All actors run concurrently and each is able to execute, or **fire**, at any time after its input tokens arrive. Input tokens can come from initial program input or as a result of earlier execution (i.e., the output of prior actor firings). The potential overall concurrency of execution is thus a function of the data dependencies that exist in the program, a degree of parallelism referred to as the **dataflow limit**.

The key observation to be made about dataflow computing that its execution is *inherently parallel* – actors function independently (asynchronously) and fire as necessary. In contrast, the von Neumann execution model involves the sequential processing of a pre-ordered set of instructions. Thus, execution is *inherently serial*. When comparing dataflow to von Neumann, a more subtle difference (yet one at the heart of the distinction between the two) to be noted is that the scheduling of instructions is determined at run-time (i.e., dynamic scheduling), whereas in a von Neumann system it occurs at compile-time (i.e., static scheduling). Figure 1 illustrates the difference between the dataflow and von Neumann approaches applied to the execution of a simple program. The program requires the multiplication of two independent additions. Under the von Neumann style of execution, the ADD operations must be executed sequentially, even though they are independent of each other, because an instruction counter schedules one instruction at a time. In contrast, since the availability of data drives the scheduling of a dataflow machine, both ADD operations can be executed as soon as their input dependencies are fulfilled.

Dataflow systems have evolved from the classic static (Dennis 1974) model to dynamic tagged token models (Arvind & Nikhil, 1990) that allowed multiple tokens per arc, to hybrid models that combine von Neumann and traditional dataflow styles of execution (Iannucci, 1988; Evripidou & Gaudiot, 1991; Gao, 1993). Other models that have been applied to digital signal processing include boolean dataflow and synchronous dataflow (Lee & Messerschmitt, 1987), resulting in architectures known as "dataflow networks". The work described in this paper is most relevant to a specific hybrid dataflow approach, known as **threaded dataflow** (Papadopoulos & Traub, 1991), which maintains a data-driven model of execution but associates instruction streams with individual threads that execute in a von Neumann fashion. It is distinct from pure von Neumann multithreading in the sense that data, not an instruction counter, remains the basis for scheduling instructions (operators). But it is also distinct from pure dataflow in the sense that execution of instruction streams is a statically scheduled sequential task, unlike the typical dynamic scheduling found in dataflow machines. As a result, threaded dataflow can also be viewed as data-driven multithreading.

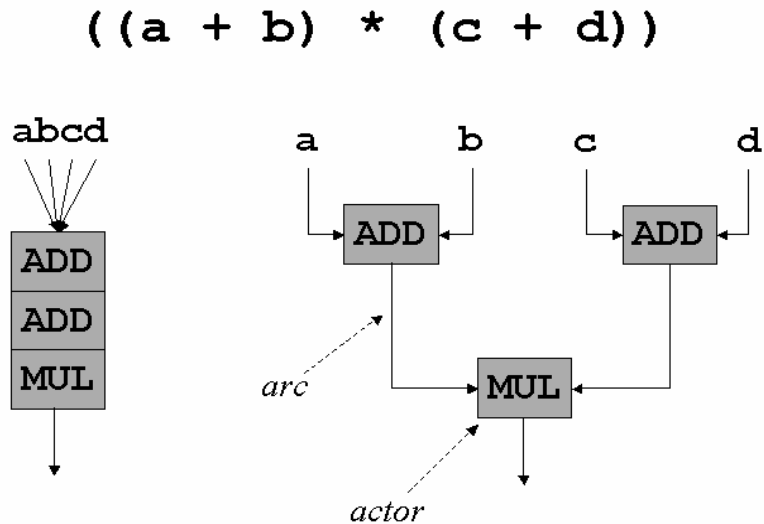

**Figure 1: Comparing von Neuman and dataflow computing**

Recent advances in processor architecture, such as the Simultaneous Multithreading (SMT) project (Tullsen, Eggers, & Levy, 1995) have demonstrated the benefits of data-driven multithreading. SMT-style processors differ from conventional CPUs (such as the Intel Pentium) by partitioning on-chip resources so that multiple threads can execute concurrently, making better use of available functional units on the same amount of chip real estate. The resulting execution reduces "vertical waste" (the wasting of cycles) that can occur when a sequence of instructions is executed using only one thread, as well as "horizontal waste" (the wasting available functional units) that can occur when executing multiple threads. To do so, the technique effectively trades instruction-level parallelism (ILP) benefits for thread-level parallelism (TLP) benefits. Instead of having a deep processor pipeline (which becomes less useful as its depth increases), SMT processors contain multiple shorter pipelines, each associated with a single thread. The result can, for highly parallel applications, substantially improve the scheduling of on-chip resources that, on conventional CPUs, would normally be starved as a result of both I/O stalls as well as thread context-switching.

The work described here applies a threaded dataflow design to a higher level of execution – the information gathering plan level. Instead of executing fine-grained instructions, we are interested in the execution of coarse-grained operators. Still, we believe that threaded dataflow is generally an efficient strategy for executing I/O-bound information gathering plans that integrate multiple remote sources because it allows coarse-grained I/O requests (such as network requests to multiple Web sources) to be automatically scheduled in parallel. Such plans are similar to other systems that maintain high degrees of concurrent network connections, such as a Web server or database system. Prior studies on such Web servers (Redstone, Eggers, & Levy, 2000) and database systems (Lo, Barroso, Eggers, Gharachorloo, Levy, & Parekh, 1998) have already shown that such systems run very efficiently on SMT-style processors; we believe the same will hold true for the execution of dataflow-style information gathering plans.

## 2.2 Web-based Information Gathering and Integration

Generic information integration systems (Chawathe et al., 1994; Arens et al., 1996; Levy et al., 1996; Genesereth et al., 1997) are concerned with the problem of allowing multiple distributed information sources to be queried as a logical whole. These systems typically deal with heterogeneous sources – in addition to traditional databases, they provide transparent access to flat files, information agents, and other structured data sources. A high-level domain model maps domain-level entities and attributes to underlying sources and the information they provide. An **information mediator** (Wiederhold 1996) is responsible for query processing, using the domain model and information about the sources to compile a query plan. In traditional databases, query processing involves three major phases: (a) parsing the query, (b) query plan generation and optimization and (c) execution. Query processing for information integration involves the same phases but builds upon traditional query plan optimization techniques by addressing cases that involve duplicate, slow, and/or unreliable information sources.

**Web-based information integration** differs from other types of information integration by focusing on the specific case where information sources are Web sites (Knoblock et al., 2001). This adds two additional challenges to the basic integration problem: (1) that of retrieving structured information (i.e., a relation) from a semi-structured source (Web pages written in HTML) and (2) querying data that is organized in a manner that facilitates human visual consumption, not necessarily in a strictly relational manner. To address the first challenge, Web

BARISH & KNOBLOCKsite **wrappers** are used to convert semi-structured HTML into structured relations, allowing Web sites to be queried as if they were databases. Wrappers take queries (such as those expressed in a query language like SQL) and process them on data extracted from a Web site, thus providing a transparent way of accessing unstructured information as if it were structured. Wrappers can be constructed manually or automatically, the latter using machine learning techniques (Knoblock, Lerman, Minton, & Muslea 2000; Kushmerick, 2000). While wrappers can be used to extract data from many Web sites, other sites are problematic because of how the data to be extracted is presented. One common case is where the Web site distributes a single logical relational answer over multiple physical Web pages, such as in the case of the online classifieds example described earlier. Automating interleaved navigation with gathering is required in such scenarios, yet it has received little attention in the literature. One approach is to extend traditional query answering for information integration systems to incorporate the capability for navigation (Friedman et al., 1999). However, such solutions mostly address the query processing phase and it remains an open issue regarding how to execute these types of information gathering plans efficiently.

A more recent technology for querying the Web is the **network query engine** (Ives et al., 1999; Hellerstein et al., 2000; Naughton et al., 2001). While these systems are, like mediators, capable of querying sets of Web sources, there has been a greater focus on the challenges of efficient query plan execution, robustness in the face of network failure or large data sets, that of processing XML data. Many network query engines rely on adaptive execution techniques, such as dynamic reordering of tuples among query plan operators (Avnur & Hellerstein 2000) and the double pipelined hash join (Ives et al., 1999), to overcome the inherent latency and unpredictable availability of Web sites.

An important aspect of network query engine research has been its focus on dataflow-style execution. Research on parallel database systems has long regarded dataflow-style query execution efficient (Grafe, 1993; Wilschut & Apers, 1993). However, when applied to the Web, dataflow-style processing can yield even greater speedups because (a) Web sources are remote, so the base latency of access is much higher than that of accessing local data and (b) Web data cannot be strategically pre-partitioned, as it can in shared-nothing architectures (DeWitt & Gray, 1992). Thus, because the average latency of Web data access is high, the parallelizing capability of dataflow-style execution is even more compelling than it is for traditional parallel database systems because the potential speedups are greater.

## 3. Motivating Example

As discussed earlier, while mediators and network query engines allow distributed Web data to be queried efficiently, they cannot handle some of the more complicated types of information gathering tasks because their query (and thus plan) languages do not support the degree of expressivity required. To better motivate our discussion, we now describe a detailed example of an information gathering problem that requires a more complex plan. Throughout the rest of this paper, we will refer to this example as we describe the details of our proposed agent plan language and execution system.

Our example involves using the Web to search for a new house to buy. Suppose that we want to use an online real estate listings site, such as Homeseekers (*http://www.homeseekers.com*), to locate houses that meet a certain set of price, location, and room constraints. In doing so, we want our query to run periodically over a medium duration of time (e.g., a few weeks) and have any new updates (i.e., new houses that meet our criteria) e-mailed to us as they are found.

To understand how to automate the gathering part of this task, let us first discuss how users would complete it manually. Figures 2a, 2b, and 2c show the user interface and result pages for

632

AN EXPRESSIVE LANGUAGE AND EFFICIENT EXECUTION SYSTEM FOR SOFTWARE AGENTS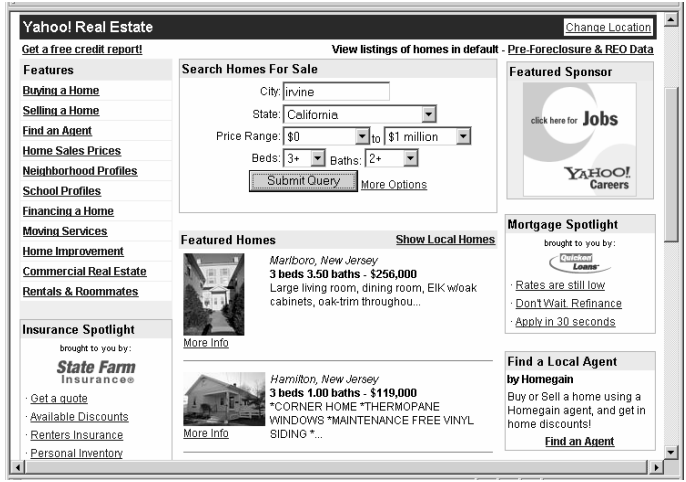

Figure 2a: Initial query form for Yahoo Real Estate

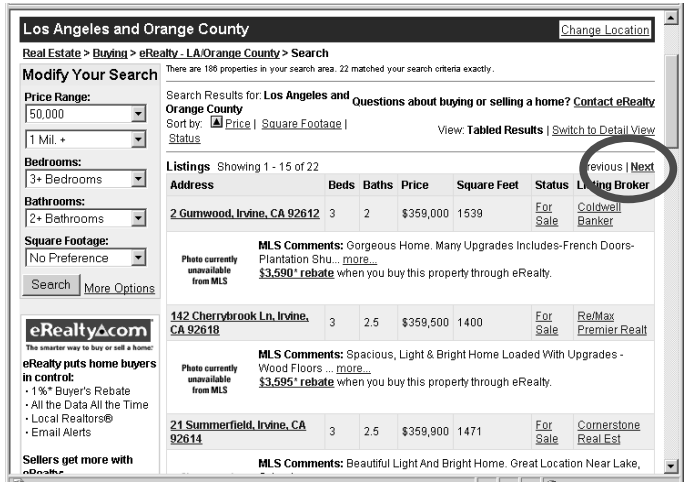

Figure 2b: Initial results from Yahoo Real Estate

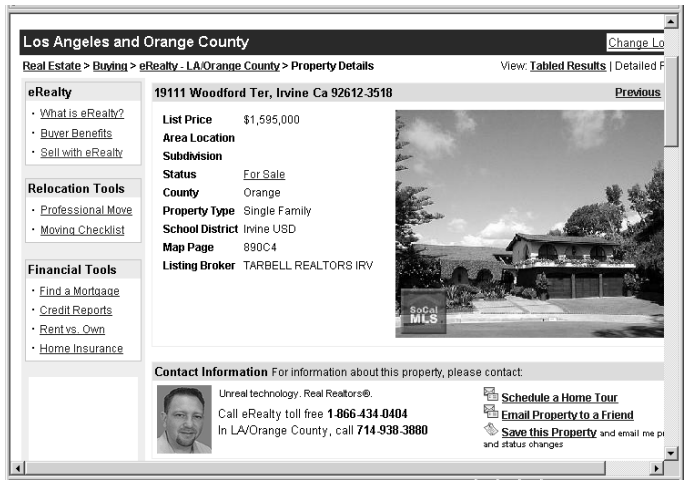

Figure 2c: Detailed result from Yahoo Real Estate

633

Homeseekers. To query for new homes, users initially fill the criteria shown in Figure 2a – specifically, they enter information that includes city, state, maximum price, etc. Once they fill in this form, they submit the query to the site and an initial set of results are returned – these are shown in Figure 2b. However, notice that this page only contains results 1 through 15 of 22. To get the remainder of the results, a "Next" link (circled in Figure 2b) must be followed to the page containing results 16 through 22. Finally, to get the details of each house, users must follow the URL link associated with *each* listing. A sample detail screen is shown in Figure 2c. The detail screen is useful because it often contains pictures and more information, such as the MLS (multiple listing services) information, about each house. In our example, the detailed page for a house must be investigated in order to identify houses that contain the number of rooms desired.

Users would then repeat the above process over a period of days, weeks, or even months. The user must both query the site periodically and keep track of new results by hand. This latter aspect can require a great deal of work – users must note which houses in each result list are new entries and identify changes (e.g., selling price updates) for houses that have been previously viewed.

As we have already discussed, it is possible to accomplish part of our task using existing Web query techniques, such as those provided by mediators and network query engines. However, notice that our task requires actions beyond gathering and filtering data. It involves periodic execution, comparison with past results, conditional execution, and asynchronous notification to the user. These are not actions that traditional Web query languages support – indeed, these actions involve more than gathering and filtering. Instead of a query plan language, what is needed is an agent plan language that supports the operators and constructs necessary to complete the task.

We can consider how such agent plans generally might look. Figure 3 shows an abstract plan for monitoring Homeseekers. As the figure shows, search criteria are used as input to generate one or more pages of house listing results. The URLs for each house from each results page are extracted and then compared against houses that already existed in a local database. New houses – those on the web page but not in the database – are subsequently queried for their details and appended to the database so that future queries can distinguish new results. During the extraction of houses from a given Homeseekers results page, the "Next" link (if any) on that page is followed and the houses on that page go through the same process. The next-link processing cycle stops when the last result page, the page without a "Next" link, has been reached. Then, after the details of the last house have been gathered, an update on the set of new houses found is e-mailed to the user.

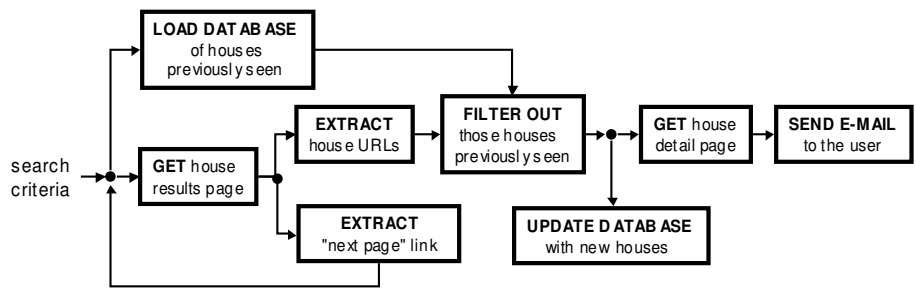

**Figure 3: Abstract plan for monitoring Yahoo Real Estate**

## 4. An Expressive Plan Language for Software Agents

In this section, we present an agent plan language that makes it possible to construct plans capable of more complicated tasks. Throughout this section, we focus on information gathering tasks, such as the Homeseekers example shown in Figure 3.

### 4.1 Plan Representation

In our language, plans are textual representations of dataflow graphs describing a set of input data, a series of operations on that data (and the intermediate results it leads to), and a set of output data. As discussed earlier, dataflow is a naturally efficient paradigm for information gathering plans. Graphs consist of a set of operator sequences (flows) where data from one operator in a given flow is iteratively processed and then flows to successive operators in the flow, eventually being merged with another flow or output from the plan.

For example, Figure 4 illustrates the dataflow graph form of a plan named *Example_plan*. It shows that the plan consists of six nodes (operators) connected with a set of edges (variables). The solid directed edges (labeled *a, b, c, d, f, and g*) represent a stream of data, while the dashed directed edge (labeled *e*) represents a signal used for synchronization purposes.

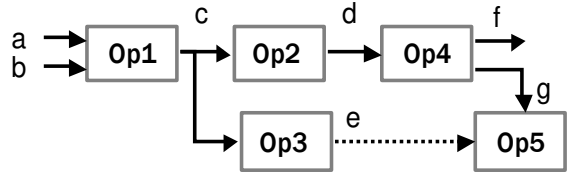

**Figure 4: Graph form of *Example_plan***

Figure 5 shows the text form of the same plan. The header consists of the name of the plan (*example_plan*), a set of input variables (*a* and *b*), and a set of output variables (*f*). The body section of the plan contains the set of operators. The set of inputs for each operator appears to the left of the colon delimiter and the set of outputs appears to the right of the delimiter. One operator (Op3) has a WAIT clause that is associated with the production of the signal indicated in Figure 4 by *e*. The ENABLE clause of a later operator (Op5) describes the consumption of that signal.

Both the graph and text forms of the example plan describe the following execution. Variables *a* and *b* are plan input variables. Together, they trigger the execution of Op1, which produces variable *c*. Op2 fires when *c* becomes available, and this leads to the output of variable *d*. Op3 fires upon the availability of *d* and produces the signal *e*. Op4 uses *d* to compute *f* (the

```
PLAN example_plan
{
  INPUT: a, b
  OUTPUT: f

  BODY
  {
    Op1 (a, b : c)
    Op2 (c : d)
    Op3 (c : )  {ENABLE: e}
    Op4 (d : f, g)
    Op5 (g : )  {WAIT : e}
  }
}
```

**Figure 5: Text form of *Example_plan***

plan output variable) and *g*. Finally, the availability of *g* and the signal *e* triggers the execution of Op5.

Note that although the body part of the text form of the plan lists operators in a linear order, this ordering does not affect when they are actually executed. Per the dataflow model of processing, operators fire whenever their individual data dependencies are fulfilled. For example, although Op3 follows Op2 in the order specified by the plan text, it actually executes at the same logical time as Op2. Also note that plan output, *f*, can be produced while the plan is still running (i.e., while Op5 is still processing).

4.1.1. FORMAL DEFINITIONS

Formally, we define the following:

**Definition 1**: *An **information gathering plan** P can be represented as a directed graph of **operators** Ops as nodes connected through a set of variables Vars that are the edges. Each plan is associated with a subset of Vars that are **plan input variables** PlanIn and another subset of variables that are **plan output variables** PlanOut. More specifically, let a plan P be represented as the tuple*

$P = <Vars, Ops, PlanIn, PlanOut>$

*where*

$Vars = \{v_1, ..., v_n\}, n > 0$
$Ops = \{Op_1, ..., Op_m\}, m > 0$
$PlanIn = \{v_{a1}, ..., v_{ax}\}, x > 0, s.t. \{v_{a1}, ..., v_{ax}\} \in Vars$
$PlanOut = \{v_{b1}, ..., v_{by}\}, y >= 0, s.t. \{v_{b1}, ..., v_{by}\} \in Vars$

**Definition 2:** *A **plan operator** Op encapsulates a **function** Func that computes a set of **operator output variables** OpOut from a set of **operator input variables** OpIn. More specifically, let each operator $Op_i$ in P be represented as the tuple*

$Op_i = <OpIn, OpOut, Func>$

*where*

$OpIn = \{v_{i1}, ..., v_{ic}\}, c > 0, s.t. \{v_{i1}, ..., v_{ic}\} \in Vars$
$OpOut = \{v_{o1}, ..., v_{og}\}, g >= 0, s.t. \{v_{o1}, ..., v_{og}\} \in Vars$
$Func$ = Function that computes $\{v_{o1}, ..., v_{og}\}$ from $\{v_{i1}, ..., v_{ic}\}$

*Furthermore, any plan $P_a$ can also be called from another plan $P_b$ as an operator. In this case, the plan $P_a$ is known as a **subplan**.*

**Definition 3:** *The schedule of execution for any operator instance $Op_i$ is described by a **firing rule** $\Psi_i$ that depends on OpIn, an optional second set of input **wait variables** OpWait, and results in the generation of OpOut and an optional second set of output **enablement variables** OpEnable. The initial firing of an operator is conditional on the availability of at least one of OpIn and all of OpWait. After the initial firing, any OpEnable variables declared are also produced. All other OpOut variables are produced in accordance with the semantics of the operator. More specifically, let us define:*

$\Psi_i (Op_i) = <OpIn, OpWait, OpOut, OpEnable>$

*where*

$OpWait = \{v_{w1}, ..., v_{wd}\}, d >= 0, s.t. \{v_{w1}, ..., v_{wd}\} \in Vars$
$OpEnable = \{v_{e1}, ..., v_{eh}\}, h >= 0, s.t. \{v_{e1}, ..., v_{eh}\} \in Vars$

Wait and enable variables are synchronization mechanisms that allow operator execution to be conditional beyond its normal set of input data variables. To understand how, let us first distinguish between a standard data variable and a synchronization variable. A standard **data variable** is one that contains information that is meant to be interpreted, or more specifically, processed by the function that an operator encapsulates. For example, *PlanIn*, *PlanOut*, *OpIn*, and *OpOut* all consist of normal data variables. A **synchronization variable** (earlier called a "signal") is one that consists of data not meant to be interpreted – rather, such variables are merely used as additional conditions to execution. Since control in dataflow systems is driven by the availability of data, synchronization variables in dataflow style plans are useful because they provide more control flow flexibility. For example, if a certain static operation should occur each time a given data flow is active, synchronization variables allow us to declare such behavior.

Definition 3 indicates that, like actors in traditional dataflow programs, operators in information gathering plans have a firing rule that describes when an operator can process its input. For example, in the dataflow computer specified by (Dennis 1974), actors can fire when their incoming arcs contain data. For the plans in the language described in this paper, the firing rule is slightly different:

> *An operator may fire upon receipt of any input variable, providing it has received all of its wait variables.*

Note that both plans and operators require at least one input because, as the firing rule implies, plans or operators without at least one input would fire continuously.

**4.2 Data Structures**

Operators process and transmit data in terms of **relations**. Each relation $R$ consists of a set of **attributes** (i.e., columns) $a_1..a_c$ and a set of zero or more **tuples** (i.e., rows) $t_1..t_r$, each tuple $t_i$ containing values $v_{i1}..v_{ic}$. We can express relations with attributes and a set of tuples containing values for each of those attributes as:

$$R (a_1, ..., a_c) = \{\{v_{11}, ..., v_{1c}\}, \{v_{21}, ..., v_{2c}\}, ..., \{v_{r1}, ..., v_{rc}\}\}$$

Each attribute of a relation can be one of five types: *char*, *number*, *date*, *relation* (embedded), or *document* (i.e., a DOM object).

Embedded relations (Schek & Scholl, 1986) within a particular relation $R_x$ are treated as opaque objects $v_{ij}$ when processed by an operator. However, when extracted, they become a separate relation $R_y$ that can be processed by the rest of the system. Embedded relations are useful in that they allow a set of values (the non-embedded objects) to be associated with an entire relation. For example, if an operator performs a COUNT function on a relation to determine the number of tuples contained in that relation, the resulting tuple emitted from the operator can consist of two attributes: (a) the embedded relation object and (b) the value equal to the number of rows in that embedded relation. Embedded relations thus allow sets to be associated with singletons, rather than forcing a join between the two. In this sense, they preserve the relationship between a particular tuple and a relation without requiring the space for an additional key or the repeating of data (as a join would require).

XML data is supported through the *document* attribute type. XML is one type of document specified by the Document Object Model (DOM). The proposed language here contains specific operators that allow DOM objects to be converted to relations, for relations to be converted to DOM objects, and for DOM objects that are XML documents to be queried in their native form using XQuery. Thus, the language supports the querying of XML documents in their native or flattened form.

**4.3 Plan Operators**

The available operators in the plan language represent a rich set of functions that can be used to address the challenges of more complex information gathering tasks, such as monitoring. Specifically, the operators support the following classes of actions:

- **data gathering**: retrieval of data from both the network and from traditional relational databases, such as Oracle or DB2.

- **data manipulation**: including standard relational data manipulation, such as Select and Join, as well as XML-style manipulations such as XQuery.
- **data storage:** the export and updating of data in traditional relational databases.
- **conditional execution**: routing of data based on its contents at run-time.
- **asynchronous notification:** communication of intermediate/periodic results through mediums/devices where transmitted data can be queued (e.g., e-mail).
- **task administration:** the dynamic scheduling or unscheduling of plans from an external task database.
- **extensibility**: the ability to embed any special type of computation (single-row or aggregate) directly into the streaming dataflow query plan

Though operators differ on their exact semantics, they do share some similarities in how they process input and generate output. In particular, there are two modes worth noting: the automatic joining of output to input (a dependent join) and the packing (embedding) and unpacking (extracting) of relations.

In information gathering plans, it is common to use data collected from one source as a basis for querying additional sources. Later, it often becomes desirable to associate the input to the source with the output it produces. However, doing this join as a separate step can be tedious because it requires the creation of another key on the existing set of data plus the cost of a join. To simplify plans and improve the efficiency of execution, many of the operators in the language perform a **dependent join** of input tuples onto the output tuples that they produce. A dependent join simply combines the contents of the input tuple with any output tuple(s) it generates, preserving the parity between the two. For example, the operator ROUND converts a floating point value in a column to its nearest whole integer value. Thus, if the input data consisted of the tuples {{Jack, 89.73}, {Jill, 98.21}} then the result after the ROUND operator executes would be of {{Jack, 89.73, 90}, {Jill, 98.21, 98}}. Without a dependent join, a primary key would need to be added (if one did not already exist) and then a separate join would have to be done after the ROUND computation. Thus, dependent joins simplify plans – they reduce the total number of operators in plan (by reducing the number of decoupled joins) and eliminate the need to ensure entity integrity prior to processing.

Another processing mode of operators involves the *packing* and *unpacking* of relations[3]. These operations are relevant in the context of embedded relations. Instead of creating and managing two distinct results (which often need to be joined later), it is cleaner and more space-efficient to perform a dependent join on the packed version of an input relation with the result output by an aggregate-type operator. For example, when using an AVERAGE operator on the input data above, the result after a dependent join with the packed form of the original relation would be: {{{Jack, 89.73}, {Jill, 98.21}}, 93.97}. Unpacking would be necessary to get at the original data. In short, embedded relations make it easy to associate aggregates with the values that led to their derivation. Packing and unpacking are useful data handling techniques that facilitate this goal.

Table 1 shows the entire set of operators in the proposed language. Some of these (such as Select and Join) have well-known semantics (Abiteboul, Hull, & Vianu, 1995) and are used in other database and information gathering systems. As a result, we will not discuss them here in any detail. However, many of the operators are new and provide the ability to express more complicated types of plans. We now focus on the purpose and mechanics of some of these other operators.

---

[3] Note: These operations are also referred to as NEST and UNNEST in database literature.

AN EXPRESSIVE LANGUAGE AND EFFICIENT EXECUTION SYSTEM FOR SOFTWARE AGENTS

| Operator | Purpose |
|---|---|
| **Wrapper** | Fetch and extract data from web sites into relations. |
| **Select** | Filters data from a relation. |
| **Project** | Filters attributes from a relation. |
| **Join** | Combines data from two relations, based on a specified condition. |
| **Union** | Performs a set union of two relations. |
| **Intersect** | Finds the intersection of two relations. |
| **Minus** | Subtracts one relation from another. |
| **Distinct** | Returns tuples unique across one or more attributes. |
| **Null** | Conditionally routes one of two streams based on existence of tuples in a third |
| **Pack** | Embeds a relation within a new relation consisting of a single tuple. |
| **Unpack** | Extracts an embedded relation from tuples of an input relation. |
| **Format** | Generates a new formatted text attribute based on tuple values. |
| **Rel2xml** | Converts a relation to an XML document. |
| **Xml2rel** | Converts an XML document to a relation. |
| **Xquery** | Queries an XML document attribute of tuples of an input relation using language specified by the Xquery standard, returning an XML document result attribute contained in the tuples of the output relation. |
| **DbImport** | Scan a table from a local database. |
| **DbQuery** | Query the schema from a local database using SQL. |
| **DbAppend** | Appends a relation to an existing table – creates the table if none exists. |
| **DbExport** | Exports a relation to a single table. |
| **DbUpdate** | Executes a SQL-style update query; no results returned. |
| **Email** | Uses SMTP to communicate an email message to a valid email address. |
| **Phone** | Sends a text message to a valid cell phone number. |
| **Fax** | Faxes data to a recipient at a valid fax number. |
| **Schedule** | Adds a task to the task database with scheduling information. |
| **Unschedule** | Removes a task from the database. |
| **Apply** | Executes a user-defined function on each tuple of a relation. |
| **Aggregate** | Executes a user-defined function on an entire relation. |

**Table 1: The complete set of operators**

4.3.1. INTERACTING WITH LOCAL DATABASES

There are two major reasons why it is useful to be able to interact with local database systems during plan execution. One reason is that the local database may contain information that we wish to integrate with other online information. A second reason has to do with the ability for the local database to act as "memory" for plans that run continuously or when a plan run at a later time needs to use the results of a plan run at an earlier time.

To address both needs, the database operators DbImport, DbQuery, DbExport, and DbAppend are provided. A common use for these operators is to implement a monitoring-style query. For example, suppose we wish to gradually collect data over a period of time, such as the collection of house data in the Homeseekers example. To accomplish this, DbImport or DbQuery can be used to bring previously queried data into a plan so that it can be compared with newly queried data (gathered by a Wrapper operator) by using any of the set-theoretic operators, such as Minus) and the result or difference can be written back to the database through DbAppend or DbExport.

4.3.2. SUPPORTING CONDITIONAL EXECUTION

Conditional execution is important for plans that need to perform different actions for data based on the run-time value of that data. To analyze and conditionally route data in a plan, the language supports the Null operator. Null acts as a switch, conditionally routing one set of data based on the status of another set of data. Null refers to the predicate "Is Null", which is a conditional that executes different actions depending on the result of the evaluation. When the data is null, action A is performed; when it is not null, action B is performed. To accomplish this in dataflow, the Null operator publishes different sets of data, one for either case.

For example, suppose it is desirable to have stock quotes automatically communicated to a user every 30 minutes. Normally, quotes should be retrieved and then e-mailed. However, if the percentage price change of any stock in the portfolio is greater than 20%, then all quotes should be sent via cell phone messaging (since such communication can be more immediate). Null would be useful in such a case because it would allow a Select condition to process the check on price changes and – if there exist tuples that match the filtering criteria – allow that data to trigger an operator that communicated those results via cell phone. Otherwise, Null would route the data to an operator that communicated in the information via e-mail. In short, Null is powerful because it is a dynamic form of conditional execution in that it can be used with other operators (like Select) to activate/deactivate flows based on the runtime content of the data.

The input and output to Null is summarized in Figure 6. The input is data to be analyzed $d$, data to be forwarded upon true (null) $dt$, and the data to be forwarded upon false $df$. If $d$ is null (i.e., contains zero tuples), then $dt$ is copied as output variable $t$. Otherwise, $df$ is copied as output $f$. For example, if $d$ contains three tuples {x1, x2, x3} and if $dt$ contains five tuples {t1, t2, t3, t4, t5} and $df$ contains two tuples {f1, f2}, then only a variable $t$ containing {t1, t2, t3, t4, t5} is output. Consumers of $f$ will never receive any data.

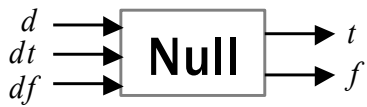

**Figure 6: The NULL operator**

4.3.3. CALLING USER-DEFINED FUNCTIONS

In designing a number of agent plans, we found that there were times when agents needed to execute some special logic (e.g., business logic) during execution. Usually, this logic did not involve relational information processing and the plan writer simply wanted to be able to code in a standard programming language (such as Java or C). For example, in some of the plans written for the Electric Elves travel agents (Ambite et al., 2002), it was necessary for the agent to send updates to users via the DARPA CoAbs Agent Grid network. In other plans, we needed to normalize the formats of date strings produced by different Web sources. Instead of expanding the operator set for each unique type of logic encountered, we developed two special operators that allowed plans to make calls to arbitrary functions written in standard programming languages. The goal of having these operators was to (a) make it easier to write plans that required special calculations or library calls, (b) encourage non-relational information processing (which could not benefit from the efficiency of dataflow style processing) to be modularized outside of the plan, and (c) to simplify plans.

The two operators, Apply and Aggregate, provide extensibility at both the tuple and relation level. Apply calls user-defined *single-row* functions on each tuple of relational data and performs a dependent join on the input tuple with its corresponding result. For example, a user-defined single-row function called SQRT might return a tuple consisting of two values: the input value and its square root. The user defined function is written in a standard programming language, such as a Java, and is executed on a per-tuple basis. Thus, this type of external function is very similar to the use of stored procedures or UDFs in commercial relational database systems.

The Aggregate operator calls user-defined *multi-row* functions and performs a dependent join on the packed form of the input and its result. For example, a COUNT function might return a relation consisting of a single tuple with two values: the first being the packed form of the input and the second being the count of the number of distinct rows in that relation. As with Apply, the user-defined multi-row function is written in a standard programming language like Java. However, in contrast to being called on a per-tuple basis, it is executed on a per-relation basis.

### 4.3.4. XML INTEGRATION

For purposes of efficiency and flexibility, it is often convenient to package or transform data to/from XML in mid-plan execution. For example, the contents of a large data set can often be described more compactly by leveraging the hierarchy of an XML document. In addition, some Web sources (such as Web services) already provide query answers in XML format. To analyze or process this data, it is often simpler and more efficient to deal with it in its native form rather than to convert it into relations, process it, and convert it back to XML. However, in other cases, a relatively small amount XML data might need to be joined with a large set of relational data.

To provide flexible XML manipulation and integration, the language supports the Rel2xml, Xml2rel, and Xquery operators. The first two convert relations to XML documents and vice-versa, using straightforward algorithms. Xml2Rel allows one to specify an "iterating" element (which map to tuples in the relation) and "attribute" elements (which map to attributes of the relation), and generates tuples that include an index referring to the order in which the original XML element was parsed. Cross product style flattening for deeper child elements is performed automatically. Rel2Xml is even more straightforward: it creates parent XML elements for each tuple and inserts attribute elements as children, in the order they appear in the relation. To allow XML to be processed in its native form, we support the Xquery operator, based on the XQuery standard (Boag et al., 2002).

The Xml2Rel, Rel2Xml, and Xquery are complementary in terms of functionality. Xml2Rel handles the basic conversion of XML to relational data, noting the order of data in the document. Rel2Xml handles the basic conversion back to XML, without regards to order – note that the nature of streaming dataflow parallelism is such that order of processing those tuples is deliberately not guaranteed. However, if the order of the XML document generated by Rel2Xml is important, Xquery can be used as a post-processing step to address that requirement. In short, both Xml2Rel and Rel2Xml focus on the simple task of converting from a relation to a document; any complex processing of XML can be accomplished through the Xquery operator.

### 4.3.5. ASYNCHRONOUS NOTIFICATION

Many continuously running plans, such as Homeseekers, do not involve interactive sessions with users. Instead, users request that a plan be run on a given schedule and expect to receive updates from the periodic execution of that plan. These updates are delivered through asynchronous means, such as e-mail, cell-phone messaging, or facsimile. To facilitate such notification, the language includes the Email, Fax, and Phone operators for communicating data via these devices.

Each of these operators works in a similar fashion. Input data received by the operator is re-formatted into a form that is suitable for transmission to the target device. The data is then transmitted: Email sends an e-mail message, Fax contacts a facsimile server with its data, and Phone routes data to cell phone capable of receiving messages.

### 4.3.6. AUTOMATIC TASK ADMINISTRATION

The overall system that accompanies the language includes a task database and a daemon process that periodically reads the task database and executes plans according to their schedule. This architecture is shown in Figure 7. Task entries consist of a plan name, a set of input to provide to that plan, and scheduling information. The latter data is represented in a format similar to the UNIX crontab entry. This format allows the minute, hour, day of the month, month, and year that a plan is supposed to be run. For example, a task entry of

```
05 08-17 1,3,5 * * homeseekers.plan
```

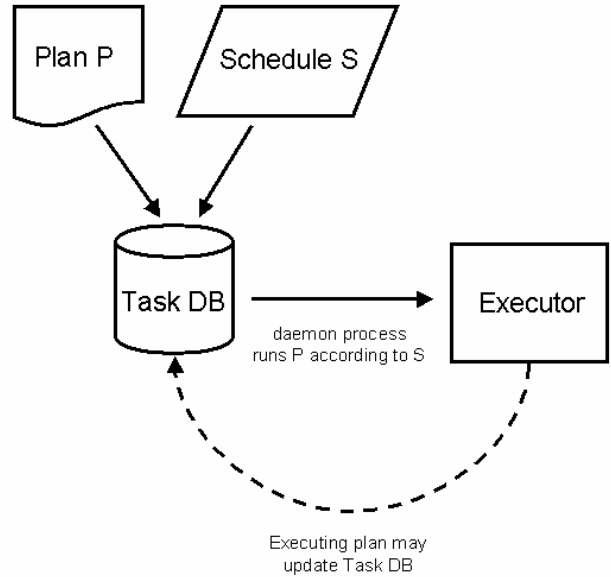

**Figure 7: Task administration process**

means: *run homeseekers.plan at five minutes after every hour between 8am and 5pm on the $1^{st}$, $3^{rd}$, and $5^{th}$ days of every month of every year*.

While tasks can be scheduled manually, the language we have developed also allows plans to automatically update the scheduling of other plans, including it. To do so, we support two special scheduling operators, Schedule and Unschedule. The former allows a plan to register a new plan to be run. It creates or updates plan schedule data in the task database. The input to Schedule consists of a plan name and a schedule description, such as the one shown above. The operator produces a single tuple of output that indicates the assigned task ID of the scheduled task.

Unschedule removes a scheduled plan from the task database. Unschedule can be used by a plan to remove itself from a monitoring activity and is often used in tandem with a notification operator. For example, a plan can monitor the set of available houses on the market for the entire month of September, send an email at the end of that month to the user containing the results, unschedule itself from execution, and then schedule a new plan (perhaps, for example, to clean up the database that stored the monitoring data). The input to Unschedule is the task ID of the scheduled plan and the output is a tuple indicating success or failure of the attempt to remove the plan from the task database.

**4.4 Subplans**

To promote reusability, modularity, and the capability for recursion, the plan language supports the notion of subplans. Recall that all plans are named, consist of a set of input and output streams, and a set of operators. If we consider that the series of operators amounts to a complex function on the input data, then plans present the same interface as do operators. In particular, using our earlier definitions, it is possible that $Op_i = P$ in that *OpIn = PlanIn*, *OpOut = PlanOut*, *OpWait* = $\varnothing$, *OpEnable* = $\varnothing$, and *Func = {$Op_1, ..., Op_n$}*. Thus, a plan can be referenced within another plan *as if it were an operator*. During execution, a subplan is called just like any other operator would – as inputs of the subplan arrive, they are executed within the body of the subplan by the operators of that subplan. For example, consider how the *Example_plan*, introduced earlier, can be referenced by another plan called *Parent_plan*. Figure 8 illustrates how the text form of *Parent_plan* treats *Example_plan* as merely another operator.

```
PLAN parent_plan
{
  INPUT: w, x
  OUTPUT: z

  BODY
  {
    Op6 (w : y)
    example_plan (x, y : z)
  }
}
```

Figure 8: Text of *parent_plan*

Subplans encourage modularity and re-use. Once written, a plan can be used as an operator in any number of future plans. Complicated manipulations of data can thus be abstracted away, making plan construction simpler and more efficient.

For example, one could develop a simple subplan called *Persistent_diff*, shown in Figure 9, that uses the existing operators DbQuery, Minus, Null, and DbAppend to take any relation, compare it to a named relation stored in a local database. This plan determines if there was an update, appends the result, and returns the difference. Many types of monitoring style plans that operate on updated results can incorporate this subplan into their existing plan. The Homeseekers plan itself could be a subplan that returns house details given a set of search parameters.

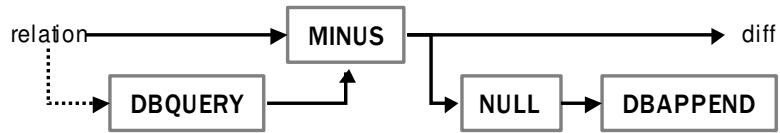

Figure 9: The *Persistent_diff* subplan

4.4.1. RECURSION

In addition to promoting modularity and re-use, subplans make another form of control flow possible: *recursion*. Recursive execution can be useful in a number of scenarios related to Web query processing. Here we describe two: reformulating data integration queries and iterating over Next Page links.

One application of recursion in THESEUS involves reformulating data integration queries. For example, a more efficient version of the Duschka's Inverse Rules algorithm (Duschka 1997) can be implemented using recursive streaming dataflow execution in THESEUS (Thakkar & Knoblock, 2003). Support for recursion in query reformulation allowed Thakkar and Knoblock to develop a system that produced more complete answers than other query reformulation algorithms, such as MiniCon (Pottinger & Levy, 2001), which do not support recursion.

Another practical use of recursion in Web data integration involves iterating over a list that is described over multiple documents. As described earlier, a number of online information gathering tasks require some sort of looping-style (repeat until) control flow. Results from a single query can be spanned across multiple Web pages. Recursion provides an elegant way to address this type of interleaved information gathering and navigation in a streaming dataflow environment.

For example, when processing results from a search engine query, an automated information gathering system needs to collect results from each page, follow the "next page" link, collect results from the next page, collect the "next page" link on that page, and so on – until it runs out of "next page" links. If we were to express this in von Neumann style programming language, a *Do...While* loop might be used accomplish this task. However, implementing these types of loops in a dataflow environment is problematic because it requires cycles within a plan. This leads to data from one loop iteration possibly colliding with data from a different iteration. In practice, loops in dataflow graphs require a fair amount of synchronization and additional operators.

Instead, this problem can be solved simply with recursion. We can use subplan reference as a means by which to repeat the same body of functionality and we can use the Null operator as the test, or exit condition. The resulting simplicity and lack of synchronization complexity makes recursion an elegant solution for addressing cases where navigation is interleaved with retrieval and when the number of iterations for looping style information gathering is not known until runtime. As an example of how recursion is used, consider the abstract plan for processing the results of a search engine query. A higher level plan called *Query_search_engine*, shown in Figure 10a, posts the initial query to the search engine and retrieves the initial results. It then processes the results with a subplan called *Gather_and_follow*, shown in Figure 10b. The search results themselves go to a Union operator and the next link is eventually used to call *Gather_and_follow* recursively. The results of this recursive call are combined at the Union operator with the first flow.

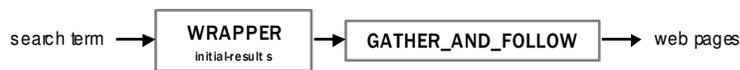

**Figure 10a: The *Query_search_engine* plan**

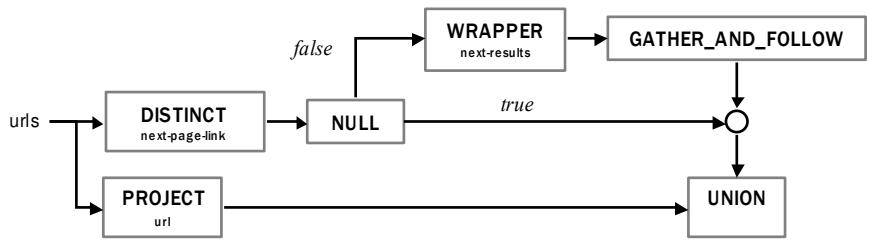

**Figure 10b: The *Gather_and_follow* recursive subplan**

4.4.2. REVISITING THE EXAMPLE

Let us now revisit the earlier house search example and see how such a plan would be expressed in the proposed plan language. Figure 11a shows one of the two plans, *Get_houses*, required to implement the abstract real estate plan in Figure 3. *Get_houses* calls the subplan *Get_urls* shown in Figure 11b, which is nearly identical to the plan *Gather_and_follow*, described above. The rest of *Get_houses* works as follows:

(a) A Wrapper operator fetches the initial set of houses and link to the next page (if any) and passes it off to the *Get_urls* recursive subplan.

(b) A Minus operator determines which houses are distinct from those previously seen; new houses are appended to the persistent store.

(c) Another Wrapper operator investigates the detail link for each house so that the full set of criteria (including picture) can be returned.

(d) Using these details, a Select operator filters out those that meet the specified search criteria.

(e) The result is e-mailed to the user.

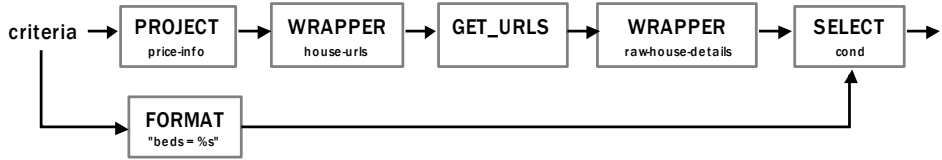

Figure 11a: The *Get_houses* plan

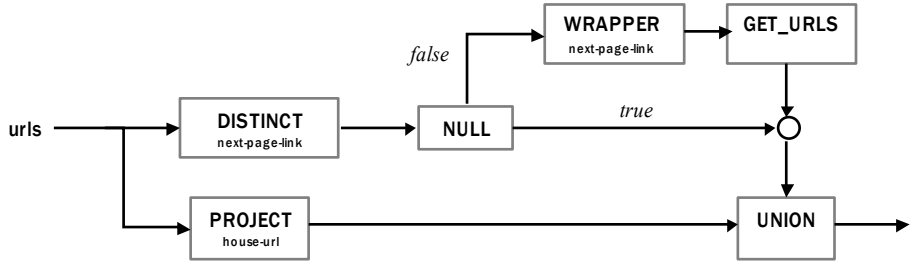

Figure 11b: The *Get_urls* recursive subplan

## 5. An Efficient Plan Execution Architecture

By definition, Web information gathering involves processing data gathered from remote sources. During the execution of an information gathering plan, it is often the case that multiple independent requests are made for different sets of remote data. Those data are then independently processed by a series of operations and then combined or output. Network latencies, bandwidth limitations, slow Web sites, and queries that yield large result sets can dramatically curtail the execution performance of information gathering plans. This is especially the case when plan operators are executed serially: any one of the issues mentioned can bottleneck the execution of an entire plan.

From an efficiency standpoint, there are two problems with standard von Neumann execution of information gathering plans. One is that it does not exploit the *independence of data flows in a common plan* in that multiple unrelated requests for remote data cannot be parallelized. The plan language we have designed addresses this problem somewhat by allowing plans to be expressed in terms of their minimal data dependencies: still, that does not dictate how those operators are actually executed.

The second efficiency problem is that von Neumann execution does not exploit the *independence of tuples in a common relation*: for example, when a large data set is being progressively retrieved from a remote source, the tuples that have already been retrieved could conceivably be operated on by successive operators in the plan. This is often reasonable, since the CPU on the local system is often under-utilized while remote data is being fetched.

To remedy both problems, we have designed a streaming dataflow execution system for software agent plans. The system allows the maximum degree of operator and data parallelism to potentially be realized at runtime, by executing multiple operators concurrently and pipelining data between operators throughout execution. Other network query engines have implemented designs that bear some similarity to what we present below. However, our discussion below extends the existing work in three ways:

- We describe the details of execution (i.e., how threads interact and how our firing rules work). With the exception of (Shah, Madden, Franklin, & Hellerstein, 2001), we have not been able to locate a similar discussion of the details of execution in these other systems.

- We present a novel thread-pooling approach to execution, where multiple threads are shared by all operators in a plan. This allows significant parallelism without exhausting resources.
- We describe how recursive streaming dataflow execution is implemented using a data coloring approach.

## 5.1 Dataflow Executor

While our plan language allows dataflow-style plans to be coded in text, it does not specify how the actual execution process works. Thus, to complement the language and to efficiently execute plans, we developed a true dataflow-style executor. The executor allows plans to realize parallelization opportunities between independent flows of data, thus enabling greater **horizontal parallelism** at runtime.

The executor functions as a virtual threaded dataflow machine. It assigns user-level threads to execute operators that are ready to fire. This type of execution is said to be "virtual dataflow" because thread creation and assignment is not done natively by the CPU, nor even in kernel space by the operating system, but by an application program (the executor) running in user space. By using threads to parallelize execution of a plan, the executor can realize better degrees of true parallelism, even on single CPU machines. This is because the use of threads reduces the impact of any I/O penalties caused by a currently executing operator. That is, multiple threads reduce the effect of vertical waste that can occur when single-threaded execution reaches an operation that blocks on I/O.

For example, consider the case where a plan containing two independent Wrapper operators is being executed on a machine with a single CPU. Suppose that both Wrapper operators have their input and can fire. Both operators will be assigned distinct threads. The single CPU will execute code that issues the network request for the first Wrapper operator, not wait for data to be returned, and finish issuing the network request for the second Wrapper operator. Thus, in a matter of microseconds, both operators will have issued their requests (which typically take on the order of hundreds of milliseconds to complete) and retrieval of the data (on the remote sites) will have been parallelized. Thus, the overall execution time will be equal to the slower of the two requests to complete. This contrasts with the execution time required for serial execution, which is equal to the sum of time required for each request.

### 5.1.1. PROMOTING AND BOUNDING PARALLELISM WITH THREAD POOLS

While using threaded dataflow has its benefits, past research in dataflow computing and operating systems has shown that there are cases when parallelism must be throttled or the overhead of thread management (i.e., the creation and destruction of threads) can be overly taxing. For example, if threads are created whenever an operator is ready, the cost to create them can add up to significant overhead. Also, if there is significant parallelism during execution, the number of threads employed might result in context switching costs that outweigh the parallelism benefits. To address both issues, we developed a thread pooling architecture that allows the executor to realize significant parallelism opportunities within fixed bounds.

At the start of plan execution, a finite number of threads are created (this number is easily adjustable through an external configuration file) and arranged in a *thread pool*. Once the threads have been created, execution begins. When data becomes available (either via input or through operator production), a thread from the pool is assigned to execute a method on the consuming operator with that data. Each time that operator produces output, it hands off the output to zero or more threads so that its consumer(s), if any, can process the output. If the pool does not contain any available threads, the output is queued in a *spillover work queue*, to be picked up later by threads as they return to the queue. This same behavior occurs for all operator input events. Thus, parallelism is both ensured by the existence of multiple threads in the pool and bounded by it – in the latter case, the degree of true parallelism during execution can never exceed the pool size. Demands on parallelism beyond the number of threads in the pool are handled by the work queue.

Figure 12 illustrates the details of how the thread pool is used by the executor at runtime. The figure shows that there are four key parts to the executor:

- **The thread pool**: This is a collection of threads ready to process the input collected in the queue. There can be a single thread pool or it can be partitioned so that certain sources have a guaranteed number of threads available to operators that query those sources. All available threads wait for new objects in the queue. Typically, contention for the queue on machines with a single CPU is not an issue (even with hundreds of threads). However, configuration options do exist for multiple work queues to be created and for the thread pool to be partitioned across queues.

- **The spillover work queue**: All data received externally and transmitted internally (i.e., as a result of operator execution) that cannot be immediately assigned to an available thread is collected in this queue. As threads return to the pool, they check if there are objects in the queue: if there are, they process them, otherwise the thread waits to be activated by future input. The queue itself is an asynchronous FIFO queue implemented as a circular buffer. When the queue is full, it grows incrementally as needed. The initial size of the queue is configurable. The structure of a queue element is described in detail below.

- **The routing table**: This data structure describes the dataflow plan graph in terms of producer/consumer operator method associations. For example, if a Select operator produces data consumed by a Project operator, the data structure that marshals output from the Select is associated with the Project input method that should consume this data. The table is computed once – prior to execution – so that the performance of operator-to-operator I/O is not impacted at runtime by repetitive lookups of consumers by producers. Instead, pre-computation allows the data structure associated with a producing method to immediately route output data to the proper set of consuming input methods.

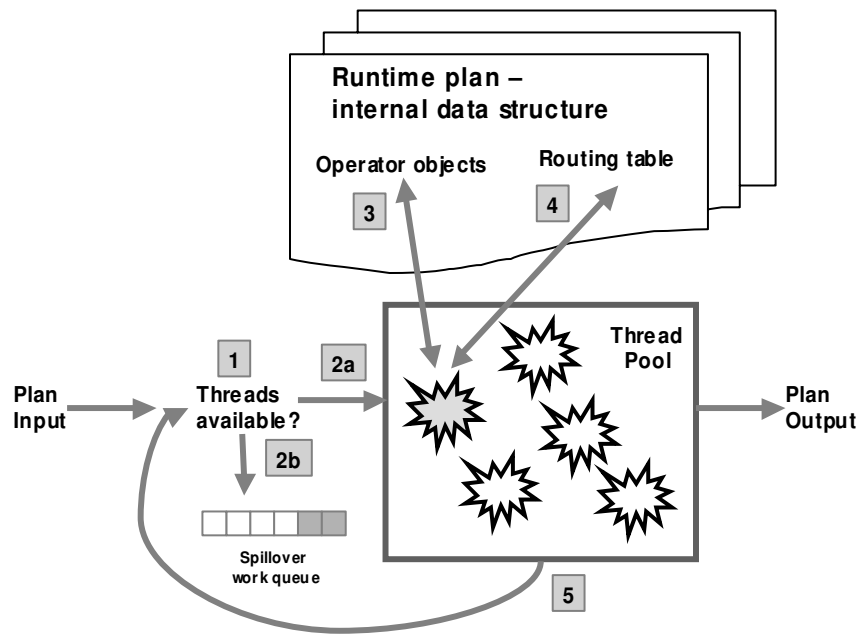

**Figure 12: Detailed design of the executor**

- **The set of operator objects**: These are the collection of operator classes (including their input/output methods and state data structures). There exists one operator object per instance in the plan.

Each queue object consists of a tuple that describes:

- the session ID
- the iteration ID
- the content (i.e., the data)
- destination operator interface (i.e., a function pointer).

The session ID is used to distinguish *independent sessions* and the iteration ID to distinguish *current call-graph level of a session*, which ensures safety during concurrent re-entrancy at runtime. These IDs provide a unique key for indexing operator state information. For example, during recursive execution, these IDs ensure that the concurrent firing of the same operator at different levels of the call graph do not co-mingle state information. Finally, the destination operator interface is the pointer to the code that the thread assigned a queue object will run.

At runtime, the system works as follows. Initial plan input arrives and is assigned to threads from the thread pool (#1 in Figure 12), one thread for each input tuple (#2a), or if no threads are available the data is added to the spillover work queue (#2b). Each assigned thread from the pool takes its queue object and, based on the description of its target, fetches the appropriate operator object so that it can execute the proper function on the data (#3). During the execution of the operator, state information from previous firings may be accessed using the (session ID, iteration ID) pair as a key. The result of an operator firing may result in output. If it does, the operator uses the routing table (#4) to determine the set of consumers of that output. It then composes new data queue objects for each consumer and hands off those objects (#5) to either an available thread in the thread pool (#2a) or deposits them to the work queue (#2b) if no threads are available. To reduce memory demands, producers only deep-copy data they produce if there are multiple consumers. Finally, operators that produce plan output data route that data out of the plan as it becomes available.

### 5.2 Data Streaming

At a logical level, each of the variables in the plan language we describe is relations. However, to provide more parallelism and thus efficiency at runtime, tuples of a common relation are **streamed** between operators. Each stream consists of stream elements (the tuples in a relation), followed by an **end of stream** (EOS) marker. Thus, when communicating a relation from producer to consumer, producing operators communicate individual tuples to consumer operators and follow the final tuple with an EOS token.

Streaming relations between operators increases the degree of **vertical parallelism** during plan execution. In revisiting the firing rule described earlier, we can further clarify it to read:

> *An operator may fire upon receipt of any input tuple, providing it has received the first tuple of all of its wait variables.*

Thus, when an operator receives a single tuple on any of its inputs, it can consume and process that tuple. Afterwards, it can potentially emit output that, in turn, can be consumed by a downstream operator or output from the plan. The resulting parallelism is "vertical" in the sense that two or more operators (e.g., one producer and one or more consumers) can concurrently operate on the same relation of data. Remote sources that return significant amounts of data can be more efficiently processed through streaming, since the operator that receives the network transmission can pass along data for processing as it becomes available and before the rest of the data has been received.

Support for any kind of streaming implies that state must be kept by operators between firings. This is because the operation being performed is logically on an entire relation, even though it physically involves each tuple of that relation. If the operator does not maintain state between firings, it cannot necessarily produce correct results. For example, consider the set-theoretic Minus operator that takes two inputs, *lhs* and *rhs*, and outputs the result of *lhs - rhs*. This operator can begin emitting output as soon as it has received the *rhs* EOS token. However, the operator must still keep track of *rhs* data until it receives the EOS from both; if not, it may emit a result that is later found to be incorrect. To see how this could happen, suppose that the order of input received by an instance of the Minus operator was:

lhs: **(Dell)**
lhs: **(Gateway)**
rhs: **(HP)**
rhs: **(Gateway)**
rhs: *EOS*
lhs: **(HP)**
lhs: *EOS*

The correct output, *lhs-rhs*, should be

lhs-rhs: **(Dell)**
lhs-rhs: *EOS*

However, this can only be achieved by waiting for the *EOS* before emitting any output and also by keeping track (i.e., maintaining state) of both inputs. For example, if only *lhs* data is retained, then the *rhs* instance of **(HP)** would not be in memory when the lhs instance of **(HP)** occurred and this tuple would be incorrectly emitted.

In summary, streaming is a technique that improves the efficiency of operator I/O by increasing the degree of vertical parallelism that is possible at runtime. By allowing producers to emit tuples as soon as possible – and by not forcing those to wait for consumers to receive them – both producers and consumers can work as fast as they are able. The main tradeoff is increased memory, for the queue required to facilitate streaming and for the state that needs to be maintained between firings.

5.2.1. RECURSIVE STREAMING: SIMPLICITY + EFFICIENCY

Streaming can complement the simplicity of many types of recursive plans with highly efficient execution. Looping in theoretical dataflow systems is non-trivial because of the desire for single-assignment and because of the need for synchronization during loop iterations. Streaming further complicates this: data from different loop iterations can collide, requiring some mechanism to color the data for each iteration. As a result, looping becomes an even more difficult challenge.

To address this problem, we use a data coloring approach. Each time that data enters a flow, it is given a session value and an iteration value (initially 0). Upon re-entrancy, the iteration value is incremented. When leaving a re-entrant module, the iteration value is decremented. If the new value is equal to 0, the flow is routed out of the recursive module; otherwise, the data flow continues to unravel until its iteration value is 0. For tail-recursive situations, the system optimizes this process and simply decrements the iteration value to 0 immediately and exits the recursive module. The two pronged data-coloring approach, which is similar to strategies used in dataflow computing literature, maintains the property of single assignment at each the level of the call graph. Streaming easily fits into this model without any other changes. As a result, many levels of the call graph can be active in parallel – effectively parallelizing the loop.

To see how this works, we return to the *Get_houses* example of Figures 11a and 11b. When the input tuple arrives, the initial page of houses is fetched. When that happens, the "Next" link is followed in parallel with the projecting of the house URLs to the Union operator and then to the Minus operator. Since the Union operator can emit results immediately, and the Minus operator will have both of its inputs, data flow continues until the next Wrapper operator, which queries the URL and extracts the details from the house. Thus, the details of the houses from the first page are queried in parallel with the following of the "Next" link, if it exists. Data from the next page is then extracted in parallel with the following of the "Next" link from this second page and so on. Meanwhile, the results from the *Get_urls* subplan (the house URLs) are streamed back to the first level of the plan, to the Union operator. They continue on through and their details are gathered in parallel.

In short, recursive streaming is a powerful capability that is made possible by the combination of the expressivity of the THESEUS plan language and efficient execution system. The result allows one to write plans that gather, extract, and process data as soon as possible – even when a logical set of results is distributed over a collection of pages (a common case on the Internet).

## 6. Experimental Results

To demonstrate the contributions of this paper, we conducted a set of experiments that highlight the increased expressivity and efficient execution supported by our plan language and execution system. Our method consists of verifying three hypotheses that are fundamental to our claims:

**Hypothesis 1:** *The streaming dataflow plan execution system ensures faster plan execution times than what is possible under von Neumann or non-streaming dataflow execution.*

**Hypothesis 2:** *The agent plan language described here supports plans that cannot be represented by the query languages of network query engines.*

**Hypothesis 3:** *The additional expressivity permitted by the plan language described here does not result in increased plan execution time.*

After a brief introduction about the implemented system used in the experiments, the rest of this section is divided into three subsections, each of which focuses on verifying each of these hypotheses.

### 6.1 The THESEUS Information Gathering System

We implemented the approach described in this paper in a system called THESEUS. THESEUS is written entirely in Java (approximately 15,000 lines of code) and thus runs on any operating system to which the Java virtual machine (JVM) has been ported. We ran the experiments described here on an Intel Pentium III 833MHz machine with 256MB RAM, running Windows 2000 Professional Edition, using the JVM and API provided by Sun Microsystems' Java Standard Edition (JSE), version 1.4. This machine was connected to the Internet via a 10Mbps Ethernet network card.

### 6.2 Hypothesis 1: Streaming Dataflow Ensures Fast Information Agents

To support our first hypothesis, we measured the efficiency of the Homeseekers information agent. Our experiments show that without the parallelism features of the plan language and execution system, agents such as Homeseekers would take significantly longer to execute.

The graphical plan for Homeseekers is the same as shown in Figures 11a and 11b. Note that this plan does not monitor Homeseekers (we will get to that in the next section), but simply gathers data from the Web site. The textual plans required for this are simply translations of Figures 11a and 11b using the plan language described in this paper. The textual form of the *Get_houses* plan is shown in Figure 13a and the textual form of the *Get_urls* plan is shown in Figure 13b.

To demonstrate the efficiency that streaming dataflow provides, we ran the Homeseekers *Get_houses* plan under three different configurations of THESEUS. The first configuration (*D-*) consisted of a thread pool with one thread – effectively preventing true multi-threaded dataflow execution and also makes streaming irrelevant. The resulting execution is thus very similar to the case where the plan had been programmed directly (without threads) using a language like Java or C++. A second THESEUS configuration (*D+S-*) used multiple threads for dataflow-style processing, but did not stream data between operators. Finally, the third configuration (*D+S+*) consisted of running THESEUS in its normal streaming dataflow mode, enabling both types of parallelism. For the *D+S-* and *D+S+* cases, the number of threads was set to 15.

Note that the configurations were only done for purposes of running these experiments. In practice, THESEUS runs in only one configuration: streaming dataflow with *n* threads in the thread pool (*n* is typically set to 10 or 20). Only if one wants to modify the number of threads in the pool does he need to alter the configuration file. This is rarely necessary.

We ran each configuration three times (interleaved, to negate any temporary benefits of network or source availability) and averaged the measurements of the three runs. The search constraints consisted of finding "houses in Irvine, CA that are priced between $500,000 and $550,000". This query returned 72 results (tuples), spread across 12 pages (6 results per page). Figure 14 shows the average performance results for these three configurations in terms of the time it took to obtain the first tuple (beginning of output) and the time it took to obtain the last tuple (end of output). A series of unpaired *t*-tests on these measurements indicates that they are

```
PLAN get_houses
{
  INPUT: criteria
  OUTPUT: filtered-house-details

  BODY
  {
    project (criteria, "price-range", price-info)
    format ("beds = %s", "beds" : bed-info)
    wrapper ("initial", price-info : result-page-info)
    get_urls (house_urls : all-house-urls)
    wrapper ("detail", all-house-urls : all-house-details)
    select (raw-house-details, bed-info : filtered-house-details)
  }
}
```

Figure 13a: Text of the *Get_houses* plan

```
PLAN get_urls
{
  INPUT: result-page-info
  OUTPUT: combined-urls

  BODY
  {
    project(result-page-info, "house-url" : curr-urls)
    distinct(result-page-info, "next-page-link" : next-status)
    null (next-status, next-status, next-status : next-page-info, next-urls)
    wrapper ("result-page", next-page-info : next-urls)
    union ( curr-urls, next-urls : combined-urls)
  }
}
```

Figure 13b: Text of the *Get_urls* plan

statistically significant at the 0.05 level.[4]

The time to first tuple is important because it shows the earliest time that data becomes available. Callers of an information agent plan are often interested in how early results come back, especially if a substantial amount of data is returned or the time between tuples is great, since it allows processing of results to begin as soon as possible. The time to last tuple is also an important metric because it is associated with the time at which all of the data has been returned. Callers of a plan that require the entire set of results, such as a caller that executes an aggregate function on the data, are thus interested in this measurement.

As Figure 14 shows, the parallelism provided by streaming dataflow has a significant impact. Typical von Neumann style execution, such as that in (*D-*), cannot not leverage opportunities for parallelism and suffers heavily from the cumulative effects of I/O delays. While the *D+S-* fares better because concurrent I/O requests can be issued in parallel, the inability to stream data throughout the plan prevents all result pages from being queried in parallel. Also, because of the lack of streaming, results obtained early during execution (i.e., the first tuple) cannot be communicated until the last tuple is ready. Note that the D+S- case reflects the performance provided if the plan had been executed by robot plan execution systems like RAPs or PRS-LITE, which support operational (horizontal) parallelism but not data (vertical) parallelism.

Finally, the *D+S+* case shows that streaming can alleviate both problems, allowing the first tuple to be output as soon as possible, while supporting the ability to query all result pages in parallel (and process the detail pages as soon as possible, in parallel). In short, Figure 14 shows that streaming dataflow is a very efficient execution paradigm for I/O-bound Web-information gathering plans that require interleaved navigation and gathering.

We also sought to compare the execution performance of the *Get_houses* plan against the performance achieved when using another type of information gathering system, such as a network query engine. However, since these systems do not support the ability to express loops or recursive information gathering, it was not possible to simply run the same plan in these other executors. To address this, we calculated the theoretical performance for a network query engine that supported streaming dataflow, but did not have the ability to loop over result pages.

To solve the type of challenge that sites like Homeseekers pose, these systems would need to gather data from one result page at a time. Note that while loops or recursion for these systems is not possible (i.e., not possible to gather data spread across a set of pages in parallel), given the type of intermediate plan language they support, they can still be used to "drill down" on the details of a particular result (i.e., gather data below a set of pages) in parallel. Thus, a network query engine could leverage its dataflow and streaming capabilities to process a single page, but could not be used to parallelize the information gathering from a set of linked result pages. Each page (and its details) would have to be processed one at a time.

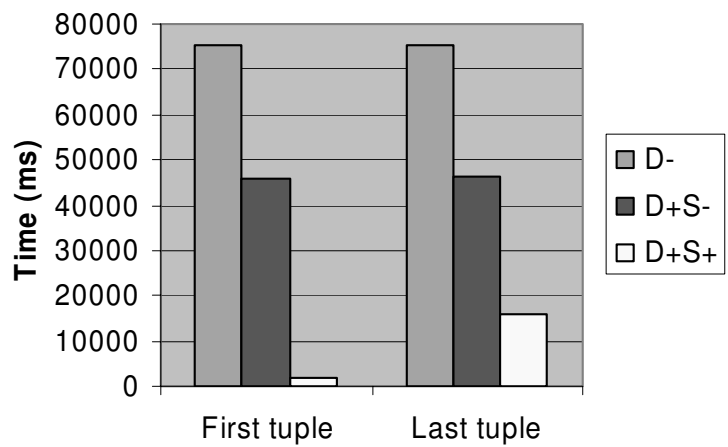

**Figure 14: Performance benefits of streaming dataflow**

---

[4] Two-tailed *P* results for the D+S vs. D+S- and D+S- vs. D- time-to-first-tuple cases were ≤ 0.0001 and 0.0024 respectively. Two-tailed P results for the D+S+ vs D+S- and D+S- vs. D- time-to-last tuple cases were 0.0001 and 0.0026, respectively:

To simulate this behavior, we used THESEUS to extract house URLs and the details one page at time, for each of the twelve pages of results we obtained in our initial query. The average time required to gather the details of all six housing results was 3204 ms. Note again that the time to retrieve the first detailed result was the same as in the THESEUS $D+S+$ case: 1852ms. If we take the time to extract all six detailed results and multiply it by the number of pages in our query (12), we get a time of last tuple equal to (3204 * 12) = 38448ms. Figure 15 shows how these results compare to the $D+S+$ case of THESEUS.

Thus, while an ad-hoc solution using a network query engine could allow the first tuple of results to be returned just as fast as in THESEUS the inability for the "Next" links to be navigated to immediately would result in less loop parallelism and, as a result, would lead to slower production of the last tuple of data. Therefore, while network query engines could be used to gather results spread across multiple hyperlinked web pages, their inability to natively support a mechanism for looping negates the potential for streaming to further parallelize the looping process.

In summary, to verify our first hypothesis, we described how the expressivity of the plan language presented enables more complex queries (like Homeseekers) to be answered efficiently. These results apply not just to Homeseekers, but to any type of site that reports a result list as a series of hyperlinked pages.

### 6.3 Hypothesis 2: Better Plan Language Expressivity

To support our second hypothesis, we investigated how the more complex task of *monitoring* Homeseekers could be accomplished using our approach versus existing Web query systems. We have previously described why monitoring in cases such as this would be useful – searching for a house is a process that requires weeks, if not months of executing the same kind of query. Thus, a corresponding information gathering plan would query Homeseekers once per day and send newly found matches to the end user over e-mail. Again, this type of problem is general – it is often desirable to be able to monitor many Internet sites that produce lists of results. However, to do so requires support for plans that are capable of expressing the monitoring task, the persistence of monitoring data, and the ability to notify users asynchronously.

The plan to monitor Homeseekers is shown in Figure 16. It is the same plan as shown in Figure 13a, but with a few additional modifications. In particular, it uses two database operators (DbImport and DbAppend) to integrate a local commercial database system for the persistence of results. This allows future queries to only return new results and stored all past results. Notice that initial DbImport is triggered by a synchronization variable. The plan also communicates new results asynchronously to users via an Email operator.

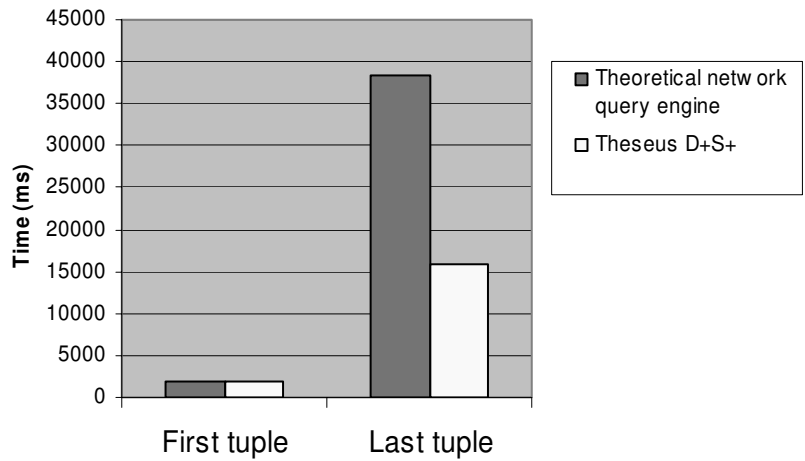

**Figure 15: Comparison against hypothetical network query engine**

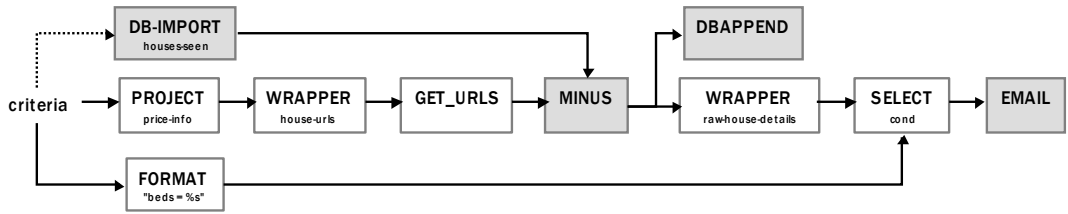

Figure 16: Modifying Homeseekers to support monitoring requirements

To measure expressivity, we consider a comparison of the plan in Figure 16 with those capable of being produced by the TELEGRAPH and NIAGARA network query engines. Our comparison focuses on TELEGRAPHCQ (Chandrasekaran, Cooper, Deshpande, Franklin, Hellerstein, Hong, Krishnamurthy, Madden, Raman, Reiss, & Shah, 2003) and NIAGARACQ (Chen, DeWitt, Tian, & Wang, 2000), both of which are modifications of their original systems to support continuous queries for the monitoring of streaming data sources. Since the TELEGRAPHCQ and NIAGARACQ query languages are very similar, we present a detailed comparison with the former and a general comparison with the latter.

Both CQ systems allow continuous Select-Project-Join (SPJ) queries to be expressed. TELEGRAPHCQ provides a SQL-like language with extensions for expressing operations on windows of streaming data. Specifically, the language allows one to express SPJ style queries over streaming data and also includes support for "for" loop constructs to allow the frequency of querying those streams. For example, to treat Homeseekers as a streaming data source and to query it once per day (for 10 days) for houses in Manhattan Beach, CA, that are less than $800,000:

```
Select street_address, num_rooms, price
  From Homeseekers
 Where price < 800000 and city = 'Manhattan Beach' and state = 'CA"
for (t=ST; t<ST+10; t++) {
  WindowIs(Homeseekers, t-1, t)
}
```

NIAGARACQ also allows more complicated operations, such as Email, to be accomplished by calling out to a function declared in a stored procedure language. The format of a NIAGARACQ query is:

```
CREATE CQ_name
XML-QL query
DO action
{START s_time} {EVERY time_interval} {EXPIRE e_time}
```

In our example, "query" would be the XML-QL equivalent of selecting house information for those that met our query criteria. The "action" part would be something similar to "MailTo:user@example.com".

Generally, both query language have the same limitations when it comes to flexible monitoring of sources, limitations that THESEUS does not have. First, there is no ability to interleave gathering of data with navigation (in fact, NIAGARACQ assumes that Homeseekers can be queried as an XML source that provides a single set of XML data). Second, there is no support for actions (like e-mail) based on differentials of data monitored over some period of time. Although both allow one to write a stored procedure that could accomplish this action, it requires a separate programming task and its execution is not as efficient as the rest of the query (this could be an issue for more complicated or more intensive CPU or I/O-bound activities per tuple). Finally, with both CQ systems, there is no way to terminate a query other than by temporal constraints.

### 6.4 Hypothesis 3: Increased Expressivity Does Not Increase Execution Time

Though we have demonstrated that THESEUS performs well on more complex information gathering tasks, it is useful to assess whether the increased expressivity in THESEUS impacts its performance on simpler tasks – in particular, ones that network query engines typically process. To do this, we explored the performance of THESEUS on a more traditional, database style query plan for online information gathering and compared it to the same type of plan executed by a network query engine.

We chose a single, common type of SPJ query that involved multiple data sources to serve as the basis for comparison. This is the canonical data integration query. We claim that understanding how THESEUS compares to a network query engine with respect to the performance of an SPJ query is at the heart of the efficiency comparison between the two types of systems. Since both types of systems execute dataflow-style plans in pipelined fashion, theoretical performance should be the same – the only expected differences would be due to implementation or environment biases (e.g., different LAN infrastructures). Nevertheless, to support our efficiency claim, we felt it was important to use a concrete SPJ query for comparison.

For our experiment, we chose to reproduce a query from the paper of another network query engine – Telegraph. To measure the performance of their partial results query processing technique, Raman and Hellerstein ran a query that gathered data from three sources and then joined them together (Raman & Hellerstein, 2002). The specific query involved gathering information on contributors to the 2000 U.S. Presidential campaign, and then combined this information with neighborhood demographic information and crime index information. Table 2 lists the sources and the data they provide. "Bulk scannable" sources are those where the data to be extracted can be read directly (i.e., exists on a static Web page or file). "Index" sources are those that provide answers based on queries via Web forms. Index sources are thus sources which require binding patterns. Table 3 shows the query that was used to evaluate the performance of TELEGRAPH.

It is important to note that Raman and Hellerstein measured the performance of the query in Table 3 under standard pipelined mode and compared this with their JuggleEddy partial results approach. We are only interested in the results of the former, as this is a measure of how well an "unoptimized" network query engine – what we call the "baseline" – gathers data when processing a traditional, database-style query. Any further optimization, such as the JuggleEddy, is complementary to the system described here. Since both types of systems rely on streaming dataflow execution consisting of tuples routed through iterative-style query operators, it would not be difficult to extend THESEUS to support this and other types of adaptive query processing techniques.

| Source | Site | Type of data |
|---|---|---|
| FEC | www.fec.gov | Bulk scannable source that provides information (including zip code) on each contributor to a candidate in the 2000 Presidential campaign. |
| Yahoo Real Estate | realestsate.yahoo.com | Index source that returns neighborhood demographic information for a particular zip code. |
| Crime | www.apbnews.com | Index source that returns crime level ratings for a particular zip code. |

**Table 2: Sources used in the FEC-Yahoo-Crime query**

| Query |
|---|
| SELECT F.Name, C.Crime, Y.income  FROM FEC as F, Crime as C, Yahoo as Y  WHERE F.zip = Y.zip and F.zip = C.zip |

Table 3: SQL query that associates crime and income statistics with political campaign contributions

We wrote a simple THESEUS plan that allowed the query in Table 3 to be executed. We used exactly the same sources, except we found that the latency of the Crime source had increased substantially, as compared to the times recorded by Raman and Hellerstein. Instead, we used another source (Yahoo Real Estate) but added an artificial delay to each tuple processed by that source, so that the new source performed similarly. Raman and Hellerstein's results show that the performance of their pipeline plan was as slow as the Crime source, and about 250ms per tuple. To match this, we added a 150ms delay to each tuple of processing for our new source, Yahoo, which was normally fetching data at about 100ms per tuple. Our results are shown in Figure 17.

The results show that THESEUS was not only able to execute the same plan at least as fast as the "baseline" TELEGRAPH plan, the non-optimized result shown in Figure 8 of the paper by Raman and Hellerstein, but THESEUS execution can be more efficient depending on the number of threads in the thread pool. For example, THESEUS-3 describes the case where the THESEUS thread pool contains 3 threads. The result from this run performs slightly worse than the TELEGRAPH baseline – such minor differences could be due to changes in source behavior or in different proximities to network sources. However, running THESEUS with more threads in the thread pool (i.e., THESEUS-6 and THESEUS-10) shows much better performance. This is because the degree of vertical parallelism demanded during execution can be better accommodated with more threads. It should be noted that the reason TELEGRAPH does not perform as well as THESEUS-6 and THESEUS-10 is likely because that system only assigned a single thread to each operator (Raman 2002). That is, THESEUS-6 and THESEUS-10 execution involves 6 and 10 concurrent threads, respectively, whereas the TELEGRAPH plan uses 3 concurrent threads.

## 7. Related Work

The language and system discussed in this paper are relevant to other efforts that focus on agent execution and the querying of Web data. To understand the work presented here in the context of these other approaches, we consider past work in software agent execution, robot agent execution, and network query engines. The first area is most relevant, as software agent systems have historically addressed expressivity issues and, in recent years, have also attempted to address some of the efficiency issues. Robot plan executors represent a slightly greater contrast that have less experience with processing large amounts of data. On the other hand, network query engines have explored large-scale remote data processing, though plan/query expressivity tends to be quite narrow.

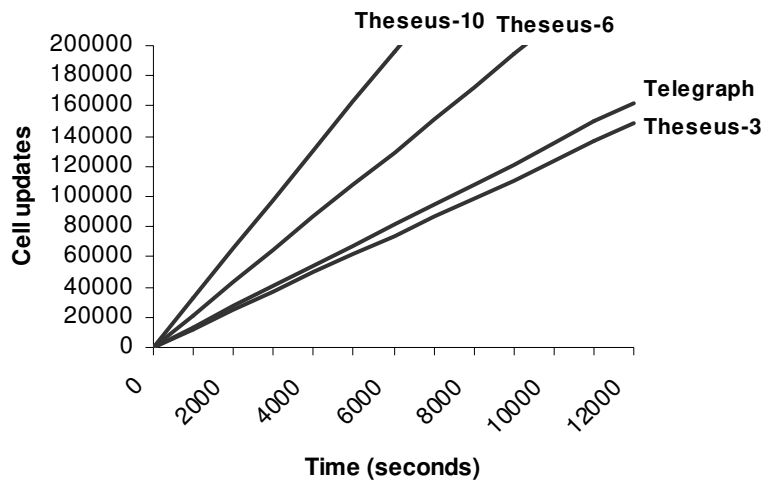

Figure 17: Comparing THESEUS to TELEGRAPH baseline (FEC-Yahoo-Crime)

**7.1 Software Agent Execution Systems**

The Internet Softbot (Eztioni & Weld, 1994) is a software agent that automates various information processing tasks, including UNIX command processing and Web information gathering. To support execution with incomplete information about the world, the system interleaves planning with execution. The XII (Golden et al., 1994) and later Puccini (Golden 1998) planners generate partially-ordered plans in which the effects of an action do not have to be known before execution – but which can be verified during execution. While the Softbot makes a clear distinction between information goals and satisfaction goals, it does not specifically address the need to efficiently nor flexibly handle the information it processed. For example, the system does not support any kind of parallel processing of information (to capitalize on the I/O-bound nature of execution). In terms of expressivity, while XII and Puccini do allow universal quantification to be expressed (i.e. iteration is possible), to do so requires that the set of what is being iterated over be known in advance. As we pointed out in an earlier example on Next Page links, this is not always the case – the set of "next" pages to be processed are only discovered by iterating through all of them in an indeterminate, *do..while* fashion. In contrast, although it does not interleave planning and execution, the system described here does support a more expressive plan language capable of handling next-link type of processing, as well as a streaming dataflow model of execution that enables efficient large scale information processing. To a great extent, contributions of both research efforts can be viewed as complementary.

Other research, such as INFOSLEUTH (Bayardo et al., 1997) has recognized the importance of concurrent task/action execution, close to the spirit of true dataflow computing. At the same time, such work has generally not investigated the impact of streaming combined with dataflow. INFOSLEUTH describes a collection of agents that, when combined and working together, present a cohesive view of data integration across multiple heterogeneous sources. INFOSLEUTH centralizes execution in its Task Execution Agent, which coordinates high-level information gathering subtasks necessary to fulfill user queries by routing appropriate queries to resources that can accommodate those queries. The Task Execution Agent is data-driven and, thus, task fulfillment proceeds in a dataflow-style manner. In addition, a multi-threading architecture supports concurrent, asynchronous communication between agents. However, the streaming component does not exist – in fact, while INFOSLEUTH intends to do large scale information processing, it specifically notes that limitations to KQML (the basis of its message transport between agents) were such that streaming was not feasible at the time of implementation. Both INFOSLEUTH and THESEUS are similar in their desire to support efficient, large-scale information processing. However, THESEUS supports streaming between operators, as well as a more expressive plan language, capable of support for more complex types of plans, including support for conditionals and recursion.

In contrast to INFOSLEUTH, BIG (Lesser, Horling, Klassner, Raja, Wagner, & Zhang, 2000) is a more general software agent that separates the components of agent planning, scheduling, and execution (among other components). BIG agents execute plans based on tasks modeled in the TÆMS modeling language. During execution, BIG reasons about resource tradeoffs and attempts to parallelize "non-local" requests (such as Web requests), at least in terms of how such actions are scheduled. In terms of expressivity, TÆMS does not include support for conditionals or looping constructs (see DECAF, below), unlike the system described in this paper. In terms of execution, BIG may perform some operations concurrently, but it does not execute in a pure dataflow manner: instead, it parallelizes only certain operations, based on whether or not they are blocking. This significantly reduces additional opportunities for dataflow-style parallelism. For example, it is not possible to parallelize CPU-bound operations (desirable on hyperthreaded processors or multi-CPU machines) nor is it possible to leverage additional I/O-bound parallelism from two different instruction flows. As an example of the latter, consider a plan that uses common input data to query a set of sources, performing different computations on the input data (e.g., to form it into a query) before each remote request. Since only I/O-bound operations are parallelized, there is no way to execute both flows simultaneously, even though both flows eventually end up I/O-bound  A second but larger difference between BIG and THESEUS is that the latter supports the capability to stream data between operators, maximizing the degree of vertical parallelism possible, while the former does not. As we have shown, better vertical parallelism during execution can yield significant performance gains.

RETSINA (Sycara et al., 2003) is a more general, multi-agent system that attempts to automate a wide range of tasks, including information processing. RETSINA is unique because it attempts to interleave not only planning and execution (as did XII in the Internet Softbot), but also information gathering. Each RETSINA agent is composed of four modules: communication, planning, scheduling, and execution monitoring. As these modules run as separate threads, communication, planning and scheduling can occur during information gathering (which is often I/O-bound). In addition, multiple actions in RETSINA can be executed concurrently, in a dataflow-style manner, through separate threads. During execution, actions communicate information between one another via provision/outcome links (Williamson, Decker, & Sycara, 1996), which are similar to the notion of operator input and output variables we have described here. While the dataflow aspect of agent execution in RETSINA is similar to that in THESEUS, its plan language is less expressive (no support for conditionals or any kind of looping, including indeterminate) and no execution support for streaming.

DECAF (Graham et al., 2003) is an extension of both the RETSINA and the TÆMS task language to support agent plans that contain *if-then-else* and looping constructs. In addition, DECAF incorporates a more advanced notion of task scheduling and views its mode of operation as more analogous to that of an operating system – for example, during execution, it is concerned with the number of I/O-bound and CPU-bound tasks at any one time, so as to optimize task scheduling. While DECAF employs a more expressive task language, closer to what is supported by THESEUS, there is no support for streaming during execution. Again, as we have shown, the benefits of increased vertical parallelism through streaming can make a significant difference when processing large amounts of data or when working with slow, remote sources, a case that is common with environments like the Web. In fact, we have shown that going beyond the dataflow limit (maximum vertical and horizontal parallelism) though techniques such as speculative execution (Barish & Knoblock, 2002; Barish & Knoblock, 2003) can yield even greater performance benefits. Streaming is not a simple feature to add to an execution system; the way operators execute must change (i.e., they become iterators), end-of-stream ordering must be managed with care, support for operator state management is needed, in addition to other related challenges.

## 7.2 Robot Agent Execution Systems

Our work on THESEUS is also related to past work on robot agent execution systems. The main similarity is the emphasis on providing both a plan language and execution system for agents. The main difference, however, is that robot agent execution systems are built primarily for robots, which act in the physical world, and lack support for some of the critical features that software agents like Web information agents require. In discussing specifics, we focus on two well-known robot agent executors: the RAP system (Firby 1994) and PRS-LITE (Myers 1996).

Both RAP and PRS-LITE offer general plan languages and execution systems that support concurrent execution of actions. Like other expressive plan languages, such as RPL (McDermott 1991), both RAP and PRS-LITE also support additional action synchronization through the WAIT-FOR clause, which triggers an action after a particular signal has been received. This is similar to the use of WAIT and ENABLE in the THESEUS plan language. PRS-LITE supports even greater expressivity, including the notion of sequencing goals, which enable conditional goals as well as parallel or sequential goal execution. For example, PRS-LITE supports the SPLIT and AND modalities as two different ways to specify parallel goal execution, the former decoupled from the parent task while the latter is more tightly coupled.

Despite the expressivity supported by RAPs and PRS-LITE, it is clear that their plan languages are primarily meant to handle the needs of robots. For example, operator execution involves the processing of signals, not streams of tuples, between operators. In contrast, both the THESEUS language and the executor are built to stream potentially large amounts of relational data. If a plan like Homeseekers was executed on RAPs or PRS-LITE, the lack of streaming would result in significantly worse performance and make poor use of available resources. This is not to say that RAPs nor PRS-LITE contain design flaws: rather, these systems simply better facilitate the needs of robots – which process small amounts of local data (such as target presence or location information) and perform actions in the physical world. In contrast, Web information agents do not act on physical objects, but software objects, such as Web sites, and need to deal with the problems associated with the unreliable remote I/O of potentially large amounts of data. Streaming is thus a critical feature for these agents, as it allows for much faster performance and for local resources, such as the CPU, to be better utilized.

Another significant difference between the language presented here and those of RAPs and PRS-LITE is the support for recursion. It is understandable that robot agent execution systems lack this feature because none of their primary tasks require such control flow. In fact, neither PRS-LITE nor RAP supports any kind of looping mechanism. In contrast, looping is often required for Web information agents, which frequently need to gather a logical set of data distributed across an indeterminate number of pages connected through "Next" page links. Recursive streaming enables high-performance looping in a dataflow environment without any kind of complicated synchronization.

It cannot be understated that features like streaming and recursion make a significant difference in terms of agent performance. For example, execution of Homeseekers without recursive streaming would fare no better than the D+S- example in Section 6, which performed much worse than the D+S+ case.

### 7.3 Network Query Engines

Network query engines such as TUKWILA (Ives et al., 1999), TELEGRAPH (Hellerstein et al., 2000) and NIAGARA (Naughton et al., 2001) have focused primarily on efficient and adaptive execution (Avnur & Hellerstein 2000; Ives et al., 2000; Shanmugasundaram et al., 2000; Raman & Hellerstein 2002), the processing of XML data (Ives et al., 2001), and continuous queries (Chen et al., 2000; Chandrasekaran et al., 2003). All of these systems take queries from users, form query plans, and execute those plans on a set of remote data sources or incoming streams. As with THESEUS, network query engines rely on streaming dataflow for the efficient, parallel processing of remote data.

The work described here differs from network query engines in two ways. The first, and most important difference, has to do with the plan language. Plans in network query engines consist mostly of relational-style operators and those required to do additional adaptive or XML-style processing. For example, TUKWILA includes a double pipelined hash join and dynamic collector operators for adaptive execution (Ives et al., 1999), and x-scan and web-join operators to facilitate the streaming of XML data as binding tuples. TELEGRAPH contains the Eddy operator (Avnur & Hellerstein 2000) for dynamic tuple routing and the SteMs operator to leverage the benefits of competing sources and access methods. NIAGARA contains the Nest operator for XML processing and other operators for managing partial results (Shanmugasundaram et al., 2000). Outside of these special operators for adaptive execution and XML processing, plans in network query engines look very similar to database style query plans. These plans are also inaccessible – users can only alter the queries that generate plans, not the plans themselves.

In contrast, the plan language presented here is more expressive and agent plans are accessible. Like network query engines, the language we have described includes relational-style operators and those for processing XML data. However, it also includes operators that support conditional execution, interaction with local databases, asynchronous notification, and user-defined single-row and aggregate functions. The plan language we developed also supports subplans for modularity, re-use, and recursive execution for looping-style information gathering. In contrast, network query engines do not support these kinds of constructs. As a result, these systems cannot represent the interleaved navigation and gathering required by tasks such as the Homeseekers example. Consider the Telegraph approach for handling Next Page links. The logic for iterating over a set of Next Page type links is located in the wrapper itself, separate from the query plan[5]. While this simplifies the wrappers somewhat (each wrapper returns all of the data for a particular site), it limits the flexibility of describing how to gather remote data. For example, if one develops a Google wrapper in Telegraph that gathers results from a search (over several pages), there is no easy way to express the requirement "stop after 10 pages" or "stop when more than 5 links from the same site are extracted". In short, since the logic for dealing with the Next Page type links has been decoupled from the plan, expressivity is limited. In addition, to build a wrapper that handles Next Page links in Telegraph, one must write a custom Java class that is referenced by the engine at runtime. In contrast, the THESEUS language can handle interleaved navigation and gathering using recursion to loop over the set of Next Page links, while streaming tuples back to the system as they are extracted, for immediate post-processing or for conditional checks (i.e., to know when to stop gathering results).

A final difference worth noting has to do with accessibility. In contrast to network query engines, plans in the language we have described are accessible to the user. Although they can be generated by query processors (Thakkar et al., 2003) and other types of applications (Tuchinda & Knoblock, 2004), just like plans produced by network query engines, they can also be constructed and modified using a text editor. This provides the ability for users to specify more complicated plans that could not otherwise be represented as a query. While some network query engines, such as NIAGARACQ (Chen et al., 2000) support some means for specifying more complicated types of actions to be associated with continuous queries, this support is not native to the system and thus it is not possible to execute complex actions in the middle of queries (such actions need to occur at certain times, for example when certain events occur). For example, NIAGARACQ requires one to specify actions in a stored procedure language, introducing a barrier (query plan to stored procedure) that does not exist in our system. Furthermore, this logic is separate from the query plan (i.e., not integrated with other query plan operators) and does not execute until some condition is met.

## 8. Conclusion and Future Work

Software agents have the potential to automate many types of tedious and time-consuming tasks that involve interactions with one or more software systems. To do so, however, requires that agent systems support plans expressive enough to capture the complexity of these tasks, while at the same time execute these plans efficiently. What is needed is a way to marry the generality of existing software agent and robot agent execution systems with the efficiency of network query engines.

In this paper, we have presented an expressive plan language and efficient approach to execution that addresses these needs. We have implemented these ideas in THESEUS and applied the system to automate many types of Web information processing tasks. The Web is a compelling domain because it is a medium which demands both agent flexibility and efficiency. While existing software agent and robot agent plan execution systems can support complex plans consisting of many different types of operators, such systems are not designed to process information as efficiently as technologies developed in the database research communities. In this paper, we have presented a plan language and execution system that combines key aspects of both agent execution systems and state-of-the-art query engines, so that software agents can efficiently accomplish complex tasks. The plan language we have described makes it possible to build agents that accomplish more complex tasks than those supported by the network query engines. Agents written using this language can then be executed as efficiently as the state-of-the-art network query engines and more efficiently than the existing agent execution systems. Beyond the work here, we have also proposed and are continuing to work on a method for speculative execution for information gathering plans (Barish & Knoblock 2002). The technique leverages machine learning techniques to analyze data produced early during execution so that accurate predictions can be made about data that will be needed later in execution (Barish & Knoblock 2003). The result is a new form of dynamic runtime parallelism that can lead to significant speedups, beyond what the dataflow limit allows.

---

[5] See the Advanced TESS Wrapper Writing section of the TESS manual, http://telegraph.cs.berkeley.edu/tess/advanced.html

We are also currently working on an Agent Wizard (Tuchinda & Knoblock, 2004), which allows the user to define agents for monitoring tasks simply by answering a set of questions about the task. The Wizard works similar to the Microsoft Excel Chart Wizard, which builds sophisticated charts by asking the user a set of simple questions. The Wizard will generate information gathering plans using the language described in this paper and schedule them for periodic execution.

## Acknowledgements

This research is based upon work supported in part by the National Science Foundation under Award No. IIS-0324955, in part by the Defense Advanced Research Projects Agency (DARPA), through the Department of the Interior, NBC, Acquisition Services Division, under Contract No. NBCHD030010, in part by the Air Force Office of Scientific Research under grant numbers F49620-01-1-0053 and FA9550-04-1-0105, in part by the United States Air Force under contract number F49620-02-C-0103, in part by a gift from the Intel Corporation, and in part by a gift from the Microsoft Corporation.

The U.S. Government is authorized to reproduce and distribute reports for Governmental purposes notwithstanding any copyright annotation thereon. The views and conclusions contained herein are those of the authors and should not be interpreted as necessarily representing the official policies or endorsements, either expressed or implied, of any of the above organizations or any person connected with them.

## References

Abiteboul, S., Hull, R. S., & Vianu, V. (1995). *Foundations of Databases*, Addison-Wesley.

Ambite, J.-L, Barish, G., Knoblock, C. A., Muslea, M., Oh, J. & Minton, S. (2002). Getting from Here to There: Interactive Planning and Agent Execution for Optimizing Travel. *Proceedings of the 14th Innovative Applications of Artificial Intelligence (IAAI-2002)*. Edmonton, Alberta, Canada.

Arens, Y, Knoblock, C. A., & Shen, W-M. (1996). "Query Reformulation for Dynamic Information Integration." *Journal of Intelligent Information Systems - Special Issue on Intelligent Information Integration* **6**(2/3): 99-130.

Arvind, Gostelow, K. P., & Plouffe, W. (1978). The Id Report: An Asynchronous Programming Language and Computing Machine, University of California, 114.

Arvind & Nikhil. R. S. (1990). "Executing a Program on the MIT Tagged-Token Dataflow Architecture." *IEEE Transactions on Computers* **39**(3): 300-318.

Avnur, R. & Hellerstein, J. M. (2000). Eddies: Continuously Adaptive Query Processing. *Proceedings of the ACM SIGMOD International Conference on Management of Data*. Dallas, TX**:** 261-272.

Barish, G. & Knoblock, C. A. (2002). Speculative Execution for Information Gathering Plans. *Proceedings of the Sixth International Conference on AI Planning and Scheduling (AIPS 2002)*. Tolouse, France: 184-193

Barish, G. & Knoblock, C. A. (2003). Learning Value Predictors for the Speculative Execution of Information Gathering Plans. *Proceedings of the 18$^{th}$ International Joint Conference on Artificial Intelligence (IJCAI 2003)*. Acapulco, Mexico: 1-8.

Boag, S., Chamberlin, D., Fernandez, M. F., Florescu, D., Robie, J., & Simeon, J. (2002). XQuery 1.0: An XML Query Language. World Wide Web Consortium, *http://www.w3.org*.

Bayardo Jr., R. J., Bohrer, W., Brice, R. S., Cichocki, A., Fowler, J., Helal, A., Kashyap, V., Ksiezyk, T., Martin, G., Nodine, M., Rashid, M., Rusinkiewicz, M., Shea, R., Unnikrishnan, C., Unruh, A., & Woelk, D. (1997). InfoSleuth: Semantic Integration of Information in Open and Dynamic Environments. *Proceedings of the ACM SIGMOD International Conference on Management of Data* (SIGMOD 1997), Tucson, AZ: 195-206

Chalupsky, H., Gil, Y., Knoblock, C. A., Lerman, K., Oh, J., Pynadath, D., Russ, T. A., & Tambe, M. (2001). Electric Elves: Applying Agent Technology to Support Human Organizations. *Proceedings of the 13th Innovative Applications of Artificial Intelligence (IAAI-2001)*. Seattle, WA.

Chandrasekaran, S., Cooper, O., Deshpande, A., Franklin, M. J., Hellerstein, J. M., Hong, W., Krishnamurthy, S., Madden, S., Raman, V., Reiss, F., & Shah, M.A. (2003). TelegraphCQ: Continuous Dataflow Processing for an Uncertain World. *Proceedings of the First Biennial Conference on Innovative Data Systems Research*. Monterey, CA.